\documentclass[letterpaper, 10 pt, journal, twoside]{IEEEtran}

\newcommand{\ignore}[1]{}

\newcommand{\bma}[1]{\left[\begin{array}{ #1}}
\newcommand{\ema}{\end{array}\right]}

\DeclareMathAlphabet{\mbf}{OT1}{ptm}{b}{n}
\newcommand{\mbs}[1]{{\boldsymbol{#1}}}

\newcommand{\mbshat}[1]{{\hat{\boldsymbol{#1}}}}

\newcommand{\mbfdot}[1]{{\dot{\mbf{#1}}}}

\newcommand{\mbfhat}[1]{{\hat{\mbf{#1}}}}
\newcommand{\mbfcheck}[1]{\ensuremath{\check{\mbf{#1}}}}

\def\fdotb{{\raisebox{-0.6ex}{ \kern0.2ex\raisebox{0.8ex}{\tiny $\hspace*{-1ex}\circ$}}}}
\def\fddotb{{\raisebox{-0.6ex}{ \kern0.2ex\raisebox{0.8ex}{\tiny $\hspace*{-1ex}\circ\circ$}}}}

\newcommand{\f}{\frac}

\newcommand{\trans}{{\ensuremath{\mathsf{T}}}} 
\newcommand{\utimes}{ {\raisebox{-0.6ex}{ \kern-1.0ex\raisebox{0.6ex}{ \small $\mathsf{v}$}}} } %
\newcommand{\beq}{\begin{equation}}
\newcommand{\eeq}{\end{equation}}
\newcommand{\bdis}{\begin{displaymath}}
\newcommand{\edis}{\end{displaymath}}
\newcommand{\beqarray}{\begin{eqnarray}}
\newcommand{\eeqarray}{\end{eqnarray}}
\newcommand{\beqarraynn}{\begin{eqnarray*}}
\newcommand{\eeqarraynn}{\end{eqnarray*}}
\newcommand{\balign}{\begin{align}}
\newcommand{\ealign}{\end{align}}
\newcommand{\balignnn}{\begin{align*}}
\newcommand{\ealignnn}{\end{align}}

\renewcommand{\theenumii}{\arabic{enumii}}
\makeatletter
\renewcommand{\p@enumii}{\theenumi.}
\makeatother

\usepackage{arydshln}
\usepackage{mathtools}

\usepackage{amsmath} 
\usepackage{amssymb}
\usepackage{float}

\usepackage{multirow}

\usepackage{capt-of}

\usepackage{cite}
\begin{document}

\title{Cascaded Filtering Using the Sigma Point Transformation (Extended Version)}

\author{Mohammed Shalaby$^{1}$, Charles Champagne Cossette$^{1}$, Jerome Le Ny$^{2}$, and James Richard Forbes$^{1}$%
\thanks{Original manuscript received: Oct. 15th, 2020; Revised Jan. 30th, 2021; Accepted Mar. 1st, 2021.}
\thanks{The original paper was recommended for publication by Editor Eric Marchand upon evaluation of the Associate Editor and Reviewers' comments.
This work was supported by FRQNT under grant 2018-PR-253646, the William Dawson Scholar program, the NSERC Discovery Grant program, and the CFI JELF program.} 
\thanks{$^{1}$M. Shalaby, C. C. Cossette, and J. R. Forbes are with the department of Mechanical Engineering, McGill University, Montreal, QC H3A 0C3, Canada. {\tt\footnotesize mohammed.shalaby@mail.mcgill.ca, charles.cossette@mail.mcgill.ca, james.richard.forbes@mcgill.ca.}}%
\thanks{$^{2}$J. Le Ny is with the department of Electrical Engineering, Polytechnique Montreal, Montreal, QC H3T 1J4, Canada. {\tt\footnotesize jerome.le-ny@polymtl.ca.}}
\thanks{Digital Object Identifier (DOI): see top of this page.}
}

\markboth{IEEE Robotics and Automation Letters. Extended Preprint Version. Accepted February, 2021}
{Shalaby \MakeLowercase{\textit{et al.}}: Cascaded Filtering Using the Sigma Point Transformation (Extended Version)} 

\maketitle

\begin{abstract}
It is often convenient to separate a state estimation task into smaller ``local" tasks, where each local estimator estimates a subset of the overall system state. However, neglecting cross-covariance terms between state estimates can result in overconfident estimates, which can ultimately degrade the accuracy of the estimator. Common cascaded filtering techniques focus on the problem of modelling cross-covariances when the local estimators share a common state vector. This letter introduces a novel cascaded and decentralized filtering approach that approximates the cross-covariances when the local estimators consider distinct state vectors. The proposed estimator is validated in simulations and in experiments on a three-dimensional attitude and position estimation problem. The proposed approach is compared to a naive cascaded filtering approach that neglects cross-covariance terms, a sigma point-based Covariance Intersection filter, and a full-state filter. In both simulations and experiments, the proposed filter outperforms the naive and the Covariance Intersection filters, while performing comparatively to the full-state filter.
\end{abstract}


\IEEEpeerreviewmaketitle

\section{Introduction} \label{sec:intro}

\IEEEPARstart{S}{tate} estimation is an integral part of technologies that use measured data to make decisions. For example, state estimation is used in time synchronization of clocks at the nanosecond level, and autonomous spatial navigation of quadcopters. For complex systems, augmenting all the required states into one monolithic state estimator can become laborious, inconvenient, or even infeasible due to computational or bandwidth limitations. Therefore, the ability to split the state estimation task into many different estimators is desirable \cite{Arambel2001, Bevly2007, Carrillo-Arce2013, Mokhtarzadeh2014, Zihajehzadeh2015, Song2018}. For example, multi-robot systems would benefit from each robot having a local estimator, and large complex systems would ideally have a set of interconnected and specialized estimators, also known as cascaded estimators. Such modularity allows the independent design, analysis, tuning, debugging, and testing of each state estimator, in addition to providing architectural clarity. More efficient computing capability is also possible as smaller state vectors can be considered in a parallel framework.

\begin{figure}
    \centering
    \includegraphics[width=\linewidth]{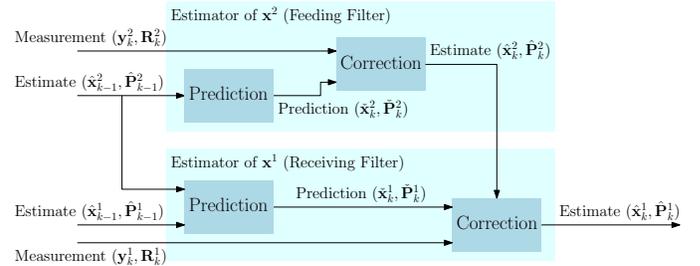}
    \caption{A block diagram of the architecture of a cascaded filter at time-step $k$. The estimator of $\mbf{x}^2$ is independent from the estimator of $\mbf{x}^1$, while the estimator of $\mbf{x}^1$ uses $\mbfhat{x}_{k-1}^2$ and $\mbfhat{x}_k^2$ as a measurement.}
    \label{fig:blockDiagram}
\end{figure}

To be able to use cascaded filtering approaches, the problem of modelling the cross-covariance between the estimates of the different filters must be addressed \cite{Arambel2001, Carlson1990, CARLSON1994, Julier1997}. Consider an attitude and heading reference system (AHRS) providing an attitude estimate to a position estimator, where the AHRS is the \emph{feeding filter} and the position estimator is the \emph{receiving filter}, as depicted in Fig.~\ref{fig:blockDiagram}. Any error in the AHRS results in an error in the position estimator, because sensors such as an accelerometer rely on attitude information when used for position estimation. Therefore, the estimation error of the attitude estimate and the position estimate are correlated. Assuming the attitude estimation error is uncorrelated with the position estimation error is equivalent to neglecting the fact that the attitude information is not entirely new, thus resulting in an overconfident estimate \cite{Mokhtarzadeh2014, Gobbini2005}. In this letter, a state estimator is said to be \emph{consistent} if its calculated error covariance does not underestimate the true error covariance. The approach of neglecting cross-covariances is common \cite{Bevly2007, Zihajehzadeh2015, Song2018, Lendek2008}, but can lead to divergence of the state estimate as a consequence of the filter inconsistency \cite{Arambel2001, Carlson1990, Julier1997}.  

To accommodate cross-covariances, a master filter is proposed in \cite{Carlson1990, CARLSON1994} named the \emph{Federated filter}, that takes as input the output of all the local, sensor-specific filters, and fuses the estimates using a least squares algorithm. This is extended in \cite{Ali2005} to nonlinear systems, by designing the master filter to be an unscented Kalman filter (UKF) \cite[Section 5.6]{sarkka2013}. A more recent modification of the Federated filter is an optimal sample-based fusion algorithm discussed in \cite{Steinbring2016}. Another popular fusion algorithm is \emph{Covariance Intersection} (CI), introduced in \cite{Julier1997}, where a convex combination of the means and covariances of different estimates of the same unknown produces a consistent fused mean and covariance. This involves the selection of weighing parameters, and in \cite{Reinhardt2014}, the authors address the optimal choice of weights in relation to state estimation of dynamical systems. A major limitation of all the discussed fusion algorithms is the focus on scenarios where the local filters share a set of common states. In \cite{Carrillo-Arce2013}, a method for computing pose estimates for a decentralized multi-robot system is presented. However, it is assumed that relative pose measurements between the robots are available to reconstruct a common shared state. Lastly, in \cite{Zhu2019}, two approaches are introduced, one based on the CI method and another that hinges on solving a linear matrix inequality problem. 

The presentation of a probabilistic approximation of the propagated cross-covariance terms when the feeding and receiving filters do not share a common set of states is a novel contribution of this letter. In particular, a novel cascaded and decentralized state estimation approach that approximates the cross-covariance terms using a sigma point transformation is introduced. Another contribution of this letter is demonstrating improved performance and consistency in simulation and experiments when compared to both an estimator that neglects cross-covariances and a sigma point-based Covariance Intersection (SPCI) estimator. The simulation and experiments consider attitude and position estimation but in a decoupled and decentralized manner. 

The remainder of this letter is organized as follows. Preliminaries are presented in Section \ref{sec:prelim}. The cascaded filtering problem is formulated in Section \ref{sec:prob_formulation}, where a probabilistic analysis of the approximate prior and posterior distributions of the receiving filter is also given. The proposed algorithm is presented in Section \ref{sec:proposed}, and is then evaluated in simulations in Section \ref{sec:sim} and in experiments in Section \ref{sec:experimental}.

\section{Preliminaries} \label{sec:prelim}

In this letter, $\mbf{x} \sim \mathcal{N}\left(\mbs{\mu}, \mbs{\Sigma}\right)$ is used to denote a Gaussian random variable $\mbf{x}$ with mean $\mbs{\mu}$ and covariance matrix $\mbs{\Sigma}$. The notation $\check{(\cdot)}$ and $\hat{(\cdot)}$ denotes the predicted and measurement-corrected state estimates, respectively. The superscripts $(\cdot)^i$ are indices, not exponents. Lastly, $\mbs{\Sigma}_{\mbf{a}\mbf{b}}$ is used to denote the cross-covariance matrix between vectors $\mbf{a}$ and $\mbf{b}$, and the following notation is used for special cross-covariance matrices,
\begin{align*}
        \mbfcheck{P}_k^i \triangleq \mbs{\Sigma}_{\mbfcheck{x}_k^i\mbfcheck{x}_k^i}, \hspace{5pt}   \mbfhat{P}_k^i \triangleq \mbs{\Sigma}_{\mbfhat{x}_k^i\mbfhat{x}_k^i},\hspace{5pt} 
        \mbfcheck{P}_{p,q}^{1,2} \triangleq \mbs{\Sigma}_{\mbfcheck{x}_p^1\mbfhat{x}_q^2}, \hspace{5pt}  \mbfhat{P}_{p,q}^{1,2} \triangleq \mbs{\Sigma}_{\mbfhat{x}_p^1\mbfhat{x}_q^2}.
\end{align*}

\subsection{The Sigma Point Transform}

The sigma point transform is a method used to approximate the nonlinear transformation of a distribution. By generating a set of sigma points from the \emph{a priori} distribution and passing them through the nonlinear function, the newly generated points are used to approximate the nonlinearly transformed distribution \cite[Section 5.5]{sarkka2013}.

Consider a nonlinear function $\mbf{z} = \mbf{h}\left( \mbf{x}, \mbf{y} \right)$, where $\mbf{x} \sim \mathcal{N}\left( \mbfhat{x}, \mbfhat{P}_x \right) \in \mathbb{R}^p$ and $\mbf{y} \sim \mathcal{N}\left( \mbfhat{y}, \mbfhat{P}_y \right) \in \mathbb{R}^q$ are mutually uncorrelated random variables. To find the distribution of $\mbf{z}$ using a sigma point transformation, the random variables are augmented into one vector $\mbf{v} \sim \mathcal{N}\left(\mbfhat{v}, \mbfhat{P}_v \right) \in \mathbb{R}^{p + q}$, where
\begin{equation}
    \mbfhat{v} = \left[ \begin{array}{c}
        \mbfhat{x}  \\
        \mbfhat{y}
    \end{array}\right], \quad \mbfhat{P}_v = \left[ \begin{array}{cc}
        \mbfhat{P}_x & \mbf{0} \\
        \mbf{0} & \mbfhat{P}_y
    \end{array} \right]. \label{prelim_spt:eq:apriori_distribution}
\end{equation}
The next step involves generating the sigma points from \eqref{prelim_spt:eq:apriori_distribution}, which can be done using different approaches. The spherical cubature rule \cite[Section 6.5]{sarkka2013} will be used throughout this letter. Define $L \triangleq \operatorname{dim} \left( \mbfhat{v} \right)$ and the Cholesky decomposition $\mbf{L}\mbf{L}^\trans \triangleq \mbfhat{P}_v$, where $\mbf{L}$ is a lower-triangular matrix. The spherical cubature rule results in a total number of $L$ sigma point pairs, where the $i^\text{th}$ pair is defined as $\mbf{s}_i \triangleq \mbfhat{v} + \sqrt{L}\operatorname{col}_i \mbf{L}$, $\mbf{s}_{i+L} \triangleq \mbfhat{v} - \sqrt{L}\operatorname{col}_i \mbf{L}$. By unstacking the $i^\text{th}$ sigma point into the two components $\mbf{x}_i$ and $\mbf{y}_i$, the $i^\text{th}$ transformed point is $\mbf{z}_i = \mbf{h}\left( \mbf{x}_i, \mbf{y}_i \right)$. The new transformed distribution can then be approximated using the $2L$ transformed points through
\begin{align}
    \mbfhat{z} &= \f{1}{2L} \sum_{i=1}^{2L} \f{1}{2L} \mbf{z}_i, \quad \mbfhat{P}_z = \f{1}{2L} \sum_{i=1}^{2L} \left( \mbf{z}_i - \mbfhat{z} \right)\left( \mbf{z}_i - \mbfhat{z} \right)^\trans.
\end{align}

\subsection{The Kalman Filter} \label{sec:kalman}

Consider a discrete-time linear system given by
\begin{align}
    \mbf{x}_k &= \mbf{A}_{k-1} \mbf{x}_{k-1} + \mbf{w}_{k-1}, \quad \mbf{w}_{k-1} \sim \mathcal{N} \left( \mbf{0}, \mbf{Q}_{k-1} \right), \\
    \mbf{y}_k &= \mbf{C}_k \mbf{x}_k + \mbs{\nu}_k, \qquad \qquad \hspace{3pt} \mbs{\nu}_{k} \sim \mathcal{N} \left( \mbf{0}, \mbf{R}_{k} \right).
\end{align}
The minimum mean square error (MMSE) estimator of $\mbf{x}_k$ given past measurements is the Kalman filter \cite[Section 4.3]{sarkka2013}, where the prediction and correction steps are given by 
\begin{align}
    \mbfcheck{x}_{k} &= \mbf{A}_{k-1} \mbfhat{x}_{k-1}, \\ \mbfcheck{P}_k &= \mbf{A}_{k-1} \mbfhat{P}_{k-1} \mbf{A}_{k-1}^\trans + \mbf{Q}_{k-1}, \\
    \mbfhat{x}_k &= \mbfcheck{x}_k + \mbf{K}_k \left( \mbf{y}_k - \mbf{C}_k \mbfcheck{x}_k \right), \\
    \mbf{K}_k &= \mbfcheck{P}_k \mbf{C}_k^\trans \left( \mbf{C}_k \mbfcheck{P}_k \mbf{C}_k^\trans + \mbf{R}_k \right)^{-1}, \\
    \mbfhat{P}_k &= \left( \mbf{1} - \mbf{K}_k \mbf{C}_k \right) \mbfcheck{P}_k, \label{eq:kalman_cov_correction}
\end{align}
with $\mbf{1}$ being the identity matrix of appropriate dimensions.

\section{Cascaded Filtering} \label{sec:prob_formulation}


Consider two discrete-time processes evolving through
\begin{align}
    \mbf{x}_k^1 &= \mbf{f}^1\left(\mbf{x}_{k-1}^1, \mbf{x}_{k-1}^2, \mbf{w}_{k-1}^1\right), \label{prelim_cascaded:eq:process_model_1} \\
    \mbf{x}_k^2 &= \mbf{f}^2\left(\mbf{x}_{k-1}^2, \mbf{w}_{k-1}^2\right), \label{prelim_cascaded:eq:process_model_2}
\end{align}
where $\mbf{x}_k^1 \in \mathbb{R}^{n_1}$ and $\mbf{x}_k^2 \in \mathbb{R}^{n_2}$ are distinct state vectors, and $\mbf{w}_{k-1}^i \sim \mathcal{N}\left( \mbf{0}, \mbf{Q}_{k-1}^i \right) \in \mathbb{R}^{\ell_i}$ represents the process noise associated with the evolution of $\mbf{x}_k^i$.
Additionally, consider two measurement signals modelled as
\begin{align}
    \mbf{y}_k^1 &= \mbf{g}^1 \left(\mbf{x}_k^1, \mbf{x}_k^2, \mbs{\nu}_k^1\right), \label{prelim_cascaded:eq:measurement_model_1} \\
     \mbf{y}_k^2 &= \mbf{g}^2 \left(\mbf{x}_k^2, \mbs{\nu}_k^2\right),  \label{prelim_cascaded:eq:measurement_model_2}
\end{align}
where $\mbf{y}_k^1 \in \mathbb{R}^{m_1}$ and $\mbf{y}_k^2 \in \mathbb{R}^{m_2}$ are distinct measurement vectors, and $\mbs{\nu}_{k}^i \sim \mathcal{N}\left( \mbf{0}, \mbf{R}_{k-1}^i \right) \in \mathbb{R}^{h_i}$ represents the measurement noise associated with $\mbf{y}_k^i$. All process and measurement noise are assumed to be mutually independent.

In the process models \eqref{prelim_cascaded:eq:process_model_1} and \eqref{prelim_cascaded:eq:process_model_2}, no inputs are considered to simplify the derivation of the proposed framework. However, this can be extended to systems with known inputs, since inputs in the receiving filter are dealt with in the standard way, and the approximation to be discussed in Section \ref{subsec:cross_cov_prop} still holds when the feeding filter has inputs.

The standard, full-state filtering approach involves designing a filter with the augmented state vector $\mbf{x}_k = [ \begin{array}{cc}
    (\mbf{x}_k^1)^\trans & (\mbf{x}_k^2)^\trans
\end{array} ]^\trans \in \mathbb{R}^{n_1 + n_2}$, using knowledge of \eqref{prelim_cascaded:eq:process_model_1}-\eqref{prelim_cascaded:eq:measurement_model_2}. Meanwhile, the cascaded filtering approach separates this problem into two filters, as shown in Fig. \ref{fig:blockDiagram}. The feeding filter outputs an approximate \emph{a posteriori} distribution of
\begin{equation}
    \mbf{x}_k^2 \big\vert \mathcal{I}_k^2 \sim \mathcal{N}\left( \mbfhat{x}_k^2, \mbfhat{P}_{k}^2 \right), \label{prelim_cascaded:eq:aposteriori_2}
\end{equation}
where $\mathcal{I}_k^i = \{ \mbfcheck{x}_0^i, \mbf{y}_{0:k}^i \}$ is an ``information" set containing a prior and measurements, $a \vert b$ denotes a random variable $a$ conditioned on $b$,  $\mbfcheck{x}^2_0 \in \mathbb{R}^{n_2}$ is the initial prediction of $\mbf{x}_k^2$, $\mbfhat{x}_k^2 \in \mathbb{R}^{n_2}$ is its estimate at time-step $k$ based on \eqref{prelim_cascaded:eq:process_model_2} and \eqref{prelim_cascaded:eq:measurement_model_2}, and $\mbfhat{P}_k^2 \in \mathbb{R}^{n_2 \times n_2}$ is the associated covariance matrix. The receiving filter uses this output alongside \eqref{prelim_cascaded:eq:process_model_1} and \eqref{prelim_cascaded:eq:measurement_model_1} to estimate the \emph{a posteriori} distribution of
\begin{equation}
    \mbf{x}_k^1 \big \vert \mbfcheck{x}^1_0, \mbf{y}_{0:k}^1, \mbfcheck{x}^2_0, \mbf{y}_{0:k}^2 \equiv \mbf{x}_k^1 \vert \mathcal{I}_k^1, \mathcal{I}_k^2 \label{prelim_cascaded:eq:aposteriori_1}
\end{equation}
without access to $\mathcal{I}_k^2$. Naive cascaded filtering \cite{Bevly2007, Zihajehzadeh2015, Song2018, Lendek2008} considers the estimate $\mbfhat{x}_k^2$ to be a measurement in the receiving filter. This breaks a requirement for consistency of the Kalman filter and its descendents, which is the conditional independence of the measurement from the state and measurement histories \cite[Property 4.2]{sarkka2013}. 

\subsection{Receiving Filter Approximate Probability Distributions}

When the \emph{a posteriori} distribution of \eqref{prelim_cascaded:eq:aposteriori_2} is known, the goal is to find a consistent estimator of \eqref{prelim_cascaded:eq:aposteriori_1} without access to the feeding filter’s process model \eqref{prelim_cascaded:eq:process_model_2}, measurement model \eqref{prelim_cascaded:eq:measurement_model_2}, or corresponding inputs and measurements. Such a solution allows the design of an $\mbf{x}^1$ estimator with minimal knowledge of the inner workings of the $\mbf{x}^2$ estimator. 

The propagation of $\mbf{x}_k^1$ is dependent on $\mbf{x}_{k-1}^2$, and mutual independence cannot be assumed. The conditional joint distribution of $\mbf{x}_k^1$ and $\mbf{x}_{k-1}^2$ is therefore assumed to be
\begin{align}
    &\left[ \begin{array}{c}
         \mbf{x}_k^1 \\
         \mbf{x}_{k-1}^2
    \end{array} \right] \Big \vert \hspace{3pt} \mathcal{I}_{k-1}^1, \mathcal{I}_{k-1}^2 \nonumber \\ &\hspace{20pt}\sim \mathcal{N} \left( \left[ \begin{array}{c}
        \mbfcheck{x}_k^1  \\
        \mbfhat{x}_{k-1}^2
    \end{array} \right], \left[ \begin{array}{cc}
        \mbfcheck{P}_k^1 & \mbfcheck{P}_{k,k-1}^{1,2}  \\
        \big(\mbfcheck{P}_{k,k-1}^{1,2}\big)^\trans & \mbfhat{P}_{k-1}^2
    \end{array} \right] \right), \label{eq:block_partitioning}
\end{align}
where $\mbfcheck{x}_k^1 = \mbf{f}^1 \left( \mbfhat{x}_{k-1}^1, \mbfhat{x}_{k-1}^2, \mbf{0} \right)$. 
This block partitioning follows from an assumption that the conditional joint distribution is Gaussian. Using \cite[Lemma A.2]{sarkka2013}, \eqref{eq:block_partitioning} gives 
\begin{align}
    &\mbf{x}_k^1 \vert \mathcal{I}_{k-1}^1, \mathcal{I}_{k-1}^2, \mbf{x}_{k-1}^2 \nonumber \\ &\hspace{20pt}\sim \mathcal{N} \Big( \mbfcheck{x}_k^1 + \mbfcheck{P}_{k,k-1}^{1,2} \big( \mbfhat{P}_{k-1}^2 \big)^{-1} \big( \mbf{x}_{k-1}^2 - \mbfhat{x}_{k-1}^2 \big),\nonumber\\&\hspace{50pt} \mbfcheck{P}_k^1 - \mbfcheck{P}_{k,k-1}^{1,2} \big(\mbfhat{P}_{k-1}^2\big)^{-1} \big(\mbfcheck{P}_{k,k-1}^{1,2}\big)^\trans\Big). \label{problem_form:eq:cond_dist_1}
\end{align}
and since $\mathcal{I}_{k-1}^2$ and $\mbf{x}_{k-1}^2$ are available only through $\mbfhat{x}_{k-1}^2 \big( \mathcal{I}_{k-1}^2 \big)$ and $\mbfhat{P}_{k-1}^2 \big( \mathcal{I}_{k-1}^2 \big)$ in the receiving filter, $\mbf{x}_{k-1}^2$ in particular is replaced in \eqref{problem_form:eq:cond_dist_1} with $\mbfhat{x}_{k-1}^2$ to obtain
\begin{align}
    &\mbf{x}_k^1 \vert \mathcal{I}_{k-1}^1, \mbfhat{x}_{k-1}^2, \mbfhat{P}_{k-1}^2 \nonumber \\&\hspace{20pt}\sim \mathcal{N} \Big( \mbfcheck{x}_k^1, \mbfcheck{P}_k^1 - \mbfcheck{P}_{k,k-1}^{1,2} \big(\mbfhat{P}_{k-1}^2\big)^{-1} \big(\mbfcheck{P}_{k,k-1}^{1,2}\big)^\trans\Big). \label{problem_form:eq:cond_dist_2}
\end{align}
In \eqref{problem_form:eq:cond_dist_2}, $\mbfhat{P}_{k-1}^2$ is known and $\mbfcheck{P}_k^1$ can be obtained using common filtering techniques. However, $\mbfcheck{P}_{k,k-1}^{1,2}$ is unknown and is required to consistently update the estimate of $\mbf{x}_k^1$.

Similarly, an approximation of the distribution of \eqref{prelim_cascaded:eq:aposteriori_1} is derived in Appendix \ref{appx:deriv_meas_update} yielding the correction step equations 
\begin{align}
    \mbfhat{x}_k^1 &= \mbfcheck{x}_k^1 + \mbf{K}_k^1\left(\mbf{y}_k^1 - \mbfcheck{y}_k^1\right) - \mbf{K}_k^{1,2} \mbf{K}_k^2\left(\mbf{y}_k^1 - \mbfcheck{y}_k^1\right), \label{eq:x1hatSimplified} \\
    \mbfhat{P}_k^1 &= \mbfcheck{P}_k^1 - \mbf{K}_k^1\mbs{\Sigma}_{\mbfcheck{x}^1_k \mbfcheck{y}^1_k}^\trans - \mbf{K}_k^{1,2}\big( \big(\mbfcheck{P}_{k,k}^{1,2}\big)^\trans - \mbf{K}_k^2\mbs{\Sigma}_{\mbfcheck{x}^1_k \mbfcheck{y}^1_k}^\trans \big), \label{eq:P1hat} \\
    \mbfhat{P}_{k,k}^{1,2} &=  \mbfcheck{P}_{k,k}^{1,2} - \mbf{K}_k^1\mbs{\Sigma}_{\mbfhat{x}^2_k \mbfcheck{y}^1_k}^\trans, \label{eq:P12hat}
\end{align}
where $\mbfcheck{y}_k^1 = \mbf{g}^1 \left( \mbfcheck{x}_k^1, \mbfhat{x}_k^2, \mbf{0} \right)$ is the predicted measurement, and
\begin{align}
    \mbf{K}_k^1 &= \mbs{\Sigma}_{\mbfcheck{x}^1_k \mbfcheck{y}^1_k}\mbs{\Sigma}_{\mbfcheck{y}^1_k \mbfcheck{y}^1_k}^{-1}, \qquad \mbf{K}_k^2 = \mbs{\Sigma}_{\mbfhat{x}^2_k \mbfcheck{y}^1_k}\mbs{\Sigma}_{\mbfcheck{y}^1_k \mbfcheck{y}^1_k}^{-1}, \hspace{-2pt}\label{eq:k1}\\
    \mbf{K}_k^{1,2} &= \left(\mbfcheck{P}_{k,k}^{1,2}- \mbf{K}_k^1\mbs{\Sigma}_{ \mbf{x}^2_k\mbf{y}^1_k}^\trans\right)\left( \mbfhat{P}_k^2-\mbf{K}_k^2\mbs{\Sigma}_{ \mbf{x}^2_k\mbf{y}^1_k}^\trans \right)^{-1}\hspace{-2pt}. \label{eq:k12simplified}
\end{align}
The form of the update equation for the cross-covariance matrix $\mbfhat{P}_{k,k}^{1,2}$ is from the joint distribution of $\mbf{x}^1$ and $\mbf{x}^2$, given in the second step of Appendix \ref{appx:deriv_meas_update}. As with the prediction step, naive cascaded filtering techniques neglect the cross-covariance components $\mbs{\Sigma}_{\mbfhat{x}^2_k \mbfcheck{y}^1_k}$ and $\mbfcheck{P}_{k,k}^{1,2}$, which appear in \eqref{eq:x1hatSimplified}-\eqref{eq:k12simplified}. Note that \eqref{eq:x1hatSimplified} and \eqref{eq:P1hat} are similar to the standard Kalman filter, except that this new form of the receiving filter also corrects the feeding filter's state $\mbf{x}^2$ locally before correcting the state $\mbf{x}^1$. The updated state of the feeding filter is never communicated back to the feeding filter, resulting in a loss of performance as compared to a full estimator.

\section{Proposed Cascaded Receiving Filter} \label{sec:proposed}

The proposed framework requires passing Gaussian distributions through nonlinear process and measurement models, which is a well-documented problem with a plethora of possible solutions. In this letter, the sigma point transformation is used as it is a concise, general approach that is suitable for handling nonlinearities. Another possible solution is a linearization-based one, which is further discussed in Appendix \ref{appx:linearization}.

In the prediction step, $\mbfhat{P}_{k-1,k-1}^{1,2}$ is used to generate sigma points, which are then propagated forward in time to approximate $\mbfcheck{P}_{k,k-1}^{1,2}$. The exact change in the cross-covariance matrix going from $\mbfcheck{P}_{k,k-1}^{1,2}$ to $\mbfcheck{P}_{k,k}^{1,2}$ is derived in Section \ref{subsec:cross_cov_prop}, and an approximation is proposed. In the correction step, $\mbfcheck{P}_{k,k}^{1,2}$ is then used to generate sigma points, which allows computing $\mbfhat{P}_{k,k}^{1,2}$.

\subsection{Prediction Step}

For prediction, the goal is to propagate the estimate $\mbfhat{x}_{k-1}^1$ forward in time to get a predicted state $\mbfcheck{x}_{k}^1$. To do so, the process model \eqref{prelim_cascaded:eq:process_model_1} and the output of the feeding estimator, $\mbfhat{x}_{k-1}^2$ and $\mbfhat{P}_{k-1}^2$, are used. Initially, $\mbfhat{P}_{0,0}^{1,2}$ is set to be the zero matrix, unless there is prior knowledge of a correlation between the initial estimates of the local estimators.

The state vectors $\mbf{x}_{k-1}^1$ and $\mbf{x}_{k-1}^2$ and the process noise $\mbf{w}_{k-1}^1$ are augmented into a vector $\mbf{v}$, with
\begin{align}
    \mbfhat{v}_{k-1} &= \left[ \begin{array}{c}
        \mbfhat{x}_{k-1}^1  \\
        \mbfhat{x}_{k-1}^2 \\
        \mbf{0}
    \end{array} \right],\label{eq:v_augment}\\ \mbfhat{P}_{\mbf{v}_{k-1}} &= \left[ \begin{array}{ccc}
        \mbfhat{P}_{k-1}^1 & \mbfhat{P}_{k-1,k-1}^{1,2} & \mbf{0} \\
        \big(\mbfhat{P}_{k-1,k-1}^{1,2}\big)^\trans & \mbfhat{P}_{k-1}^2 & \mbf{0} \\
        \mbf{0} & \mbf{0} & \mbf{Q}_{k-1}^1
    \end{array} \right]. \label{eq:P_v}
\end{align}
Let $L \triangleq \operatorname{dim}\left( \mbfhat{v}_{k-1} \right)$, and $\mbf{L}\mbf{L}^\trans \triangleq \mbfhat{P}_{\mbf{v}_{k-1}}$. Using the spherical cubature rule \cite{sarkka2013}, define the $2L$ sigma points $\mbf{s}$ to be
\begin{align*}
    \mbf{s}_{k-1}^i &\triangleq \mbfhat{v}_{k-1} + \sqrt{L} \operatorname{col}_i\mbf{L}, \quad \mbf{s}_{k-1}^{i+L} \triangleq \mbfhat{v}_{k-1} - \sqrt{L} \operatorname{col}_i\mbf{L}.
\end{align*}
By unstacking each sigma point into 
\begin{equation}
    \mbf{s}_{k-1}^i = \big[ \begin{array}{ccc}
        \big(\mbfhat{x}_{k-1,i}^1\big)^\trans &
        \big(\mbfhat{x}_{k-1,i}^2\big)^\trans &
        \big(\mbf{w}_{k-1,i}^1\big)^\trans
    \end{array} \big]^\trans,
\end{equation}
the sigma points are propagated through \eqref{prelim_cascaded:eq:process_model_1},
\begin{equation}
    \mbfcheck{x}_{k,i}^1 = \mbf{f}^1\left( \mbfhat{x}_{k-1,i}^1, \mbfhat{x}_{k-1,i}^2, \mbf{w}_{k-1,i}^1 \right).
\end{equation}
Hence, the statistics of the propagated sigma points are
        \begin{align}
            &\hspace{-7pt}\mbfcheck{x}_k^1 = \f{1}{2L}\sum_{i=1}^{2L} \mbfcheck{x}_{k,i}^1, \quad
            \mbfcheck{P}_k^1 = \f{1}{2L}\sum_{i=1}^{2L} \left( \mbfcheck{x}_{k,i}^1 - \mbfcheck{x}_k^1 \right)\left( \mbfcheck{x}_{k,i}^1 - \mbfcheck{x}_k^1 \right)^\trans, \nonumber \\
            &\hspace{15pt}\mbfcheck{P}_{k,k-1}^{1,2} = \f{1}{2L} \sum_{i=1}^{2L} \left( \mbfcheck{x}_{k,i}^1 - \mbfcheck{x}_k^1 \right)\left( \mbfhat{x}_{k-1,i}^2 - \mbfhat{x}_{k-1}^2 \right)^\trans. \label{eq:cross_covariance_time_update} 
        \end{align}

\subsection{Correction Step}

For the correction step, the goal is to correct the predicted state $\mbfcheck{x}_k^1$ using the measurement $\mbf{y}_k^1$, to obtain $\mbfhat{x}_k^1$. As $\mbf{y}_k^1$ is a function of $\mbf{x}_k^2$, the output of the feeding estimator $\mbfhat{x}_k^2$ and $\mbfhat{P}_k^2$ is required. As before, the state vectors $\mbf{x}_{k}^1$ and $\mbf{x}_{k}^2$ and the measurement noise $\mbs{\nu}_{k}^1$ are augmented into a vector $\mbf{u}$, with
\begin{equation}
    \mbfhat{u}_k = \left[ \begin{array}{c}
        \mbfcheck{x}_{k}^1  \\
        \mbfhat{x}_{k}^2 \\
        \mbf{0}
    \end{array} \right], \quad \mbfhat{P}_{\mbf{u}_k}= \left[ \begin{array}{ccc}
        \mbfcheck{P}_{k}^1 & \mbfcheck{P}_{k,k}^{1,2} & \mbf{0} \\
        \big(\mbfcheck{P}_{k,k}^{1,2}\big)^\trans & \mbfhat{P}_{k}^2 & \mbf{0} \\
        \mbf{0} & \mbf{0} & \mbf{R}_{k}^1
    \end{array} \right]. \label{eq:P_u}
\end{equation}
The problem of finding $\mbfcheck{P}_{k,k}^{1,2}$ in \eqref{eq:P_u} from \eqref{eq:cross_covariance_time_update} is addressed in Section \ref{subsec:cross_cov_prop}.
Let $M \triangleq \operatorname{dim}\left( \mbfhat{u}_k \right)$, and $\mbf{M}\mbf{M}^\trans \triangleq \mbfhat{P}_{\mbf{u}_k}$. Consequently, using the spherical cubature rule, define the $2M$ sigma points $\mbf{p}$ to be
\begin{align*}
    \mbf{p}_{k}^i &\triangleq \mbfhat{u}_k + \sqrt{M} \operatorname{col}_i\mbf{M}, \quad \mbf{p}_{k}^{i+L} \triangleq \mbfhat{u}_k - \sqrt{M} \operatorname{col}_i\mbf{M}.
\end{align*}
    By unstacking each sigma point into 
    \begin{equation}
        \mbf{p}_{k}^i = \big[ \begin{array}{ccc}
            \big(\mbfcheck{x}_{k,i}^1\big)^\trans  &
            \big(\mbfhat{x}_{k,i}^2\big)^\trans &
            \big(\mbs{\nu}_{k,i}^1\big)^\trans
        \end{array} \big]^\trans,
    \end{equation}
    the nonlinear transformation of the sigma points as per the measurement model is
    \begin{equation}
        \mbfcheck{y}_{k,i}^1 = \mbf{g}^1\left( \mbfcheck{x}_{k,i}^1, \mbfhat{x}_{k,i}^2, \mbs{\nu}_{k,i}^1 \right).
    \end{equation}
Hence, the statistics of the propagated sigma points are
\begin{align}
    \mbfcheck{y}_{k}^1 &= \f{1}{2M} \sum_{i=1}^{2M} \mbfcheck{y}_{k,i}^1,\\
    \mbs{\Sigma}_{\mbfcheck{x}_k^1\mbfcheck{y}_k^1} &= \f{1}{2M} \sum_{i=1}^{2M} \left( \mbfcheck{x}_{k,i}^1 - \mbfcheck{x}_k^1 \right) \left( \mbfcheck{y}_{k,i}^1 - \mbfcheck{y}_{k}^1 \right)^\trans, \\
    \mbs{\Sigma}_{\mbfhat{x}_k^2\mbfcheck{y}_k^1} &= \f{1}{2M} \sum_{i=1}^{2M} \left( \mbfhat{x}_{k,i}^2 - \mbfhat{x}_k^2 \right) \left( \mbfcheck{y}_{k,i}^1 - \mbfcheck{y}_{k}^1 \right)^\trans, \\
    \mbs{\Sigma}_{\mbfcheck{y}_k^1\mbfcheck{y}_k^1} &= \f{1}{2M} \sum_{i=1}^{2M} \left( \mbfcheck{y}_{k,i}^1 - \mbfcheck{y}_{k}^1 \right) \left( \mbfcheck{y}_{k,i}^1 - \mbfcheck{y}_{k}^1 \right)^\trans.
\end{align}
The filter equations given in \eqref{eq:x1hatSimplified}-\eqref{eq:k12simplified} can then be used to obtain the approximated distribution of the corrected state $\mbfhat{x}_k^1$ and its cross-covariance matrix with the state estimate $\mbfhat{x}_k^2$.

\subsection{Approximating the Effect of the Feeding Filter on the Cross-Covariance} \label{subsec:cross_cov_prop}

The cross-covariance matrix $\mbfcheck{P}_{k,k-1}^{1,2}$ is approximated using sigma points in the prediction step. However, $\mbfcheck{P}_{k,k}^{1,2}$ is needed to generate the sigma points for the correction step. If knowledge of the process and measurement models of the feeding filter is available, $\mbfcheck{P}_{k,k-1}^{1,2}$ can be propagated to $\mbfcheck{P}_{k,k}^{1,2}$ analytically. To elucidate this, consider the linear system 
\begin{align}
    \mbf{x}_k^1 &= \mbf{A}_{k-1}^1 \mbf{x}_{k-1}^1 + \mbf{B}_{k-1}^1 \mbf{x}_{k-1}^2 + \mbf{w}_{k-1}^1, \label{eq:process_model_linear1}\\
    \mbf{x}_k^2 &= \mbf{A}_{k-1}^2 \mbf{x}_{k-1}^2 + \mbf{w}_{k-1}^2, \\
    \mbf{y}_k^1 &= \mbf{C}_{k}^1\mbf{x}_k^1 + \mbf{D}_{k-1}^1 \mbf{x}_k^2 + \mbs{\nu}_k^1, \quad
    \mbf{y}_k^2 = \mbf{C}_{k}^2\mbf{x}_k^2 + \mbs{\nu}_k^2, \label{eq:models_linear2}
\end{align}
where the state vectors and noise parameters follow the same notation as in \eqref{prelim_cascaded:eq:process_model_1}-\eqref{prelim_cascaded:eq:measurement_model_2}. Moreover, consider for the moment that each of the process and measurement models in \eqref{eq:process_model_linear1}-\eqref{eq:models_linear2} are known. Let $\mbf{G}_k^2 \in \mathbb{R}^{n_2 \times m_2}$ denote the Kalman gain of the estimator of $\mbf{x}_k^2$ as per Section \ref{sec:kalman}, then
\begin{align}
    \mbfcheck{P}_{k,k}^{1,2} &= \mathbb{E} \left[ \left( \mbf{x}_k^1 - \mbfcheck{x}_k^1 \right) \left( \mbf{x}_{k}^2 - \mbfhat{x}_{k}^2 \right)^\trans \big\vert \hspace{3pt} \mathcal{I}_{k-1}^1, \mathcal{I}_{k}^2 \right] \nonumber \\
    &= \mathbb{E} \Big[ \big( \mbf{x}_k^1 - \mbfcheck{x}_k^1 \big) \big( \mbf{A}_{k-1}^2 \mbf{x}_{k-1}^2 + \mbf{w}_{k-1}^2 - \mbf{A}_{k-1}^2 \mbfhat{x}_{k-1}^2 \nonumber \\ &\hspace{25pt} - \mbf{G}_k^2 \mbf{C}_k^2 \left( \mbf{A}_{k-1}^2 \mbf{x}_{k-1}^2 + \mbf{w}_{k-1}^2 \right) - \mbf{G}_k^2 \mbs{\nu}_{k}^2 \nonumber \\ &\hspace{25pt} + \mbf{G}_k^2 \mbf{C}_k^2 \mbf{A}_{k-1}^2 \mbfhat{x}_{k-1}^2 \big)^\trans \big\vert \hspace{3pt} \mathcal{I}_{k-1}^1, \mathcal{I}_{k}^2 \Big] \nonumber \\
    &= \mathbb{E} \Big[ \left( \mbf{x}_k^1 - \mbfcheck{x}_k^1 \right) \left( \mbf{x}_{k-1}^2 - \mbfhat{x}_{k-1}^2 \right)^\trans \big\vert \hspace{3pt} \mathcal{I}_{k-1}^1, \mathcal{I}_{k}^2 \Big]  \nonumber \\
    &\hspace{85pt} \times \left(\mbf{A}_{k-1}^{2} - \mbf{G}_k^2 \mbf{C}_k^2 \mbf{A}_{k-1}^{2}\right)^\trans \nonumber \\
    &= \mbfcheck{P}_{k,k-1}^{1,2} \mbs{\Psi}_k, \label{eq:cross_cov_propagate}
\end{align}
where the assumptions $\mathbb{E} \big[ \left( \mbf{x}_k^1 - \mbfcheck{x}_k^1 \right) \left(\mbf{w}_{k-1}^2\right)^\trans \big\vert \hspace{3pt} \mathcal{I}_{k-1}^1, \mathcal{I}_{k}^2 \big] = \mbf{0}$ and $\mathbb{E} \big[ \left( \mbf{x}_k^1 - \mbfcheck{x}_k^1 \right) \left(\mbs{\nu}_{k}^2\right)^\trans \big\vert \hspace{3pt} \mathcal{I}_{k-1}^1, \mathcal{I}_{k}^2 \big] = \mbf{0}$ have been used, and $\mbs{\Psi}_k \triangleq \left( \left( \mbf{1} - \mbf{G}_k^2 \mbf{C}_k^2 \right) \mbf{A}_{k-1}^2 \right)^\trans$. 
Since, however, the estimator of $\mbf{x}^1$ does not have access to the process and measurement models of the feeding filter, computing \eqref{eq:cross_cov_propagate} exactly is not possible. If cooperation from the feeding filter is at all possible, the most accurate solution is for the feeding filter to share the matrix $\mbs{\Psi}_k$ at every time-step, in addition to the estimated distribution of $\mbf{x}^2$. When this is not possible, however, an alternative is to approximate $\mbs{\Psi}_k$. Using \eqref{eq:kalman_cov_correction}, $\mbs{\Psi}_k$ can be rewritten as 
\begin{equation}
    \mbs{\Psi}_k = \left(\mbfhat{P}_k^2 \left(\mbfcheck{P}_k^2\right)^{-1} \mbf{A}_{k-1}^2 \right)^\trans.
\end{equation}
Once the feeding filter reaches a steady state, 
$\mbfhat{P}_k^2 \left(\mbfcheck{P}_k^2\right)^{-1} \approx \mbf{1}$, yielding the approximation $\mbshat{\Psi}_k = \left(\mbf{A}_{k-1}^2\right)^\trans$. Hence, with knowledge of what $\mbshat{\Psi}$ should be if access to the process model \eqref{prelim_cascaded:eq:process_model_2} is available, the user can form an educated guess of the Jacobian $\mbf{A}^2$. Since the state vector $\mbf{x}^2$ of the feeding filter is known, the user can reconstruct an approximate (i.e., lower fidelity) process model for the states of the feeding filter, and use the Jacobian of this approximate process model as an approximation of the true Jacobian of the process model \eqref{prelim_cascaded:eq:process_model_2} used in the feeding filter. 

\begin{figure*}
	\centering
	\begin{minipage}{0.4\textwidth}
		\centering
		\includegraphics[trim=0.9cm 0.1cm 1cm 0cm, clip=true,width=\columnwidth]{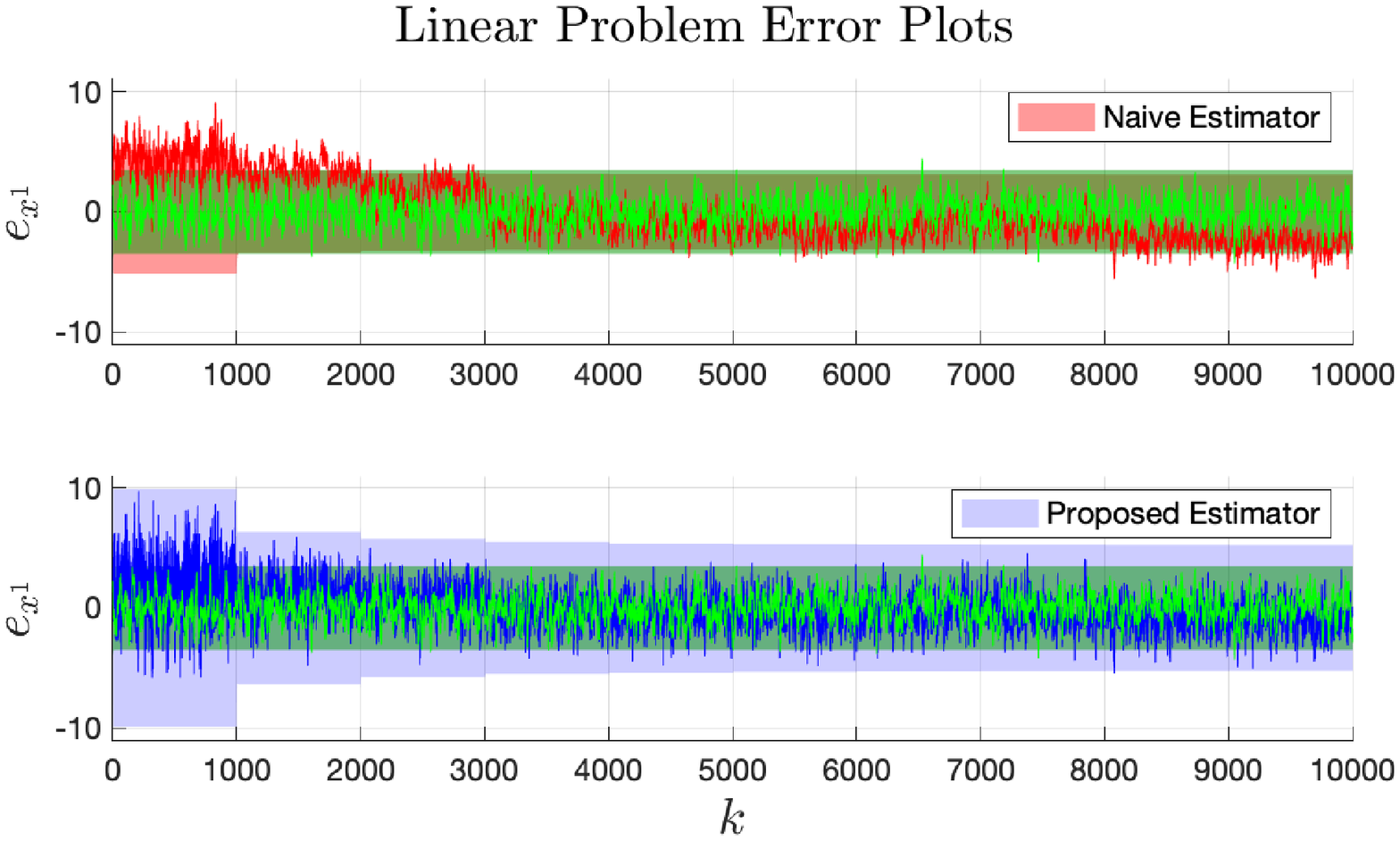}%
		\caption{Single run of the naive and proposed estimators on the linear system. Shaded regions correspond to $\pm 3 \sigma$ bounds. Red, blue, and green correspond to naive, proposed, and full estimators, respectively. After $k=1000$, the $\pm3\sigma$ bound of the naive and full estimator almost fully overlap.\vspace{10pt}}
		\label{fig:SR_linear}
	\end{minipage}\hspace{8pt}
	\begin{minipage}{0.3\textwidth}
    	\centering
    	\vspace{-15pt}
    	\includegraphics[trim=0.9cm 0.1cm 0.9cm 0cm, clip=true,width=\columnwidth]{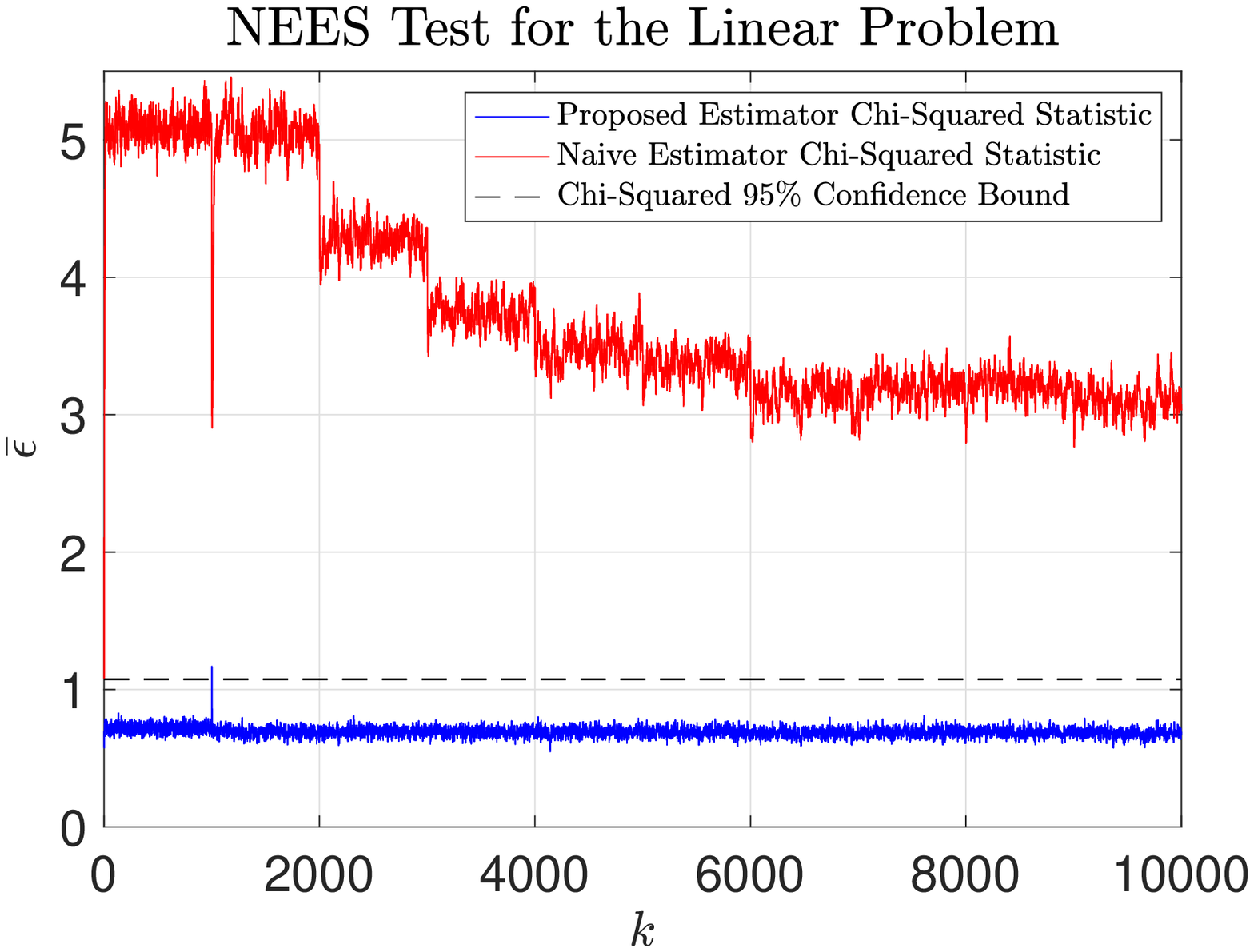}%
    	\caption{The NEES test for 1000 Monte Carlo trials on the linear system, showing the consistency evaluation of the cascaded estimators.}
    	\label{fig:MC_NEES_linear}
	\end{minipage}\hspace{8pt}
	\begin{minipage}{0.25\textwidth}
	\centering
	    \vspace{-15pt}
	    \captionof{table}{Simulation parameters for the nonlinear system.\vspace{1pt}}
        \label{tab:sim_params_nonlinear}
        \footnotesize
        \renewcommand{\arraystretch}{1.2}
        \begin{tabular}{|c|c|c|}
            \hline
            \bfseries & Specification & \bfseries Value\\
            \hline
            & Accel. (m/s$^2$) & 0.10 \\
            & Gyro. (rad/s) & 0.0032 \\
            & Mag. ($\mu$F) & 2.00 \\
            & Pos. meas. (m) & 0.22 \\
            & Initial pos. (m) & 0.45 \\
            & Initial vel. (m/s) & 0.45 \\
            \multirow{-7}{*}{\rotatebox[origin=c]{90}{Std. Deviation}} & Initial att. (rad) & 0.22 \\
            \hline
            & IMU (Hz) & 100 \\
            \multirow{-2}{*}{\rotatebox[origin=c]{90}{Rate}}& Pos. meas. (Hz) & 50 \\
            \hline
        \end{tabular}
        \normalsize
	\end{minipage}
\end{figure*}

One possible issue with over-estimating $\mbfcheck{P}^{1,2}_{k,k}$ using this approximation is that $\mbfhat{P}_{\mbf{u}_k}$ in \eqref{eq:P_u} can become indefinite. Therefore, at every iteration, the definiteness of $\mbfhat{P}_{\mbf{u}_k}$ must be checked, and the cross-covariances are to be deflated using a scalar parameter until the definiteness test passes.

\section{Simulation Results} \label{sec:sim}

The main criteria for evaluating the proposed cascaded filter is its consistency, which will be done using the normalized estimation error squared (NEES) test on Monte Carlo runs. The NEES test involves computing a chi-squared statistic $\bar{\epsilon}$ using the error trajectory and corresponding covariance of multiple trials. If the statistic is below a certain threshold, the hypothesis that the estimator is consistent cannot be rejected with 95\% confidence \cite{barshalom2002}. The $\pm3\sigma$ bound test is also considered, and the root-mean-squared-error (RMSE) is computed to evaluate the performance of the estimators. 

\subsection{Linear System} \label{sec:linear_system}

\begin{figure*}
	\begin{minipage}{0.57\textwidth}
		\centering
		\centering
        \includegraphics[trim=0.9cm 0.4cm 0.5cm 1cm, clip=true,width=0.8\columnwidth]{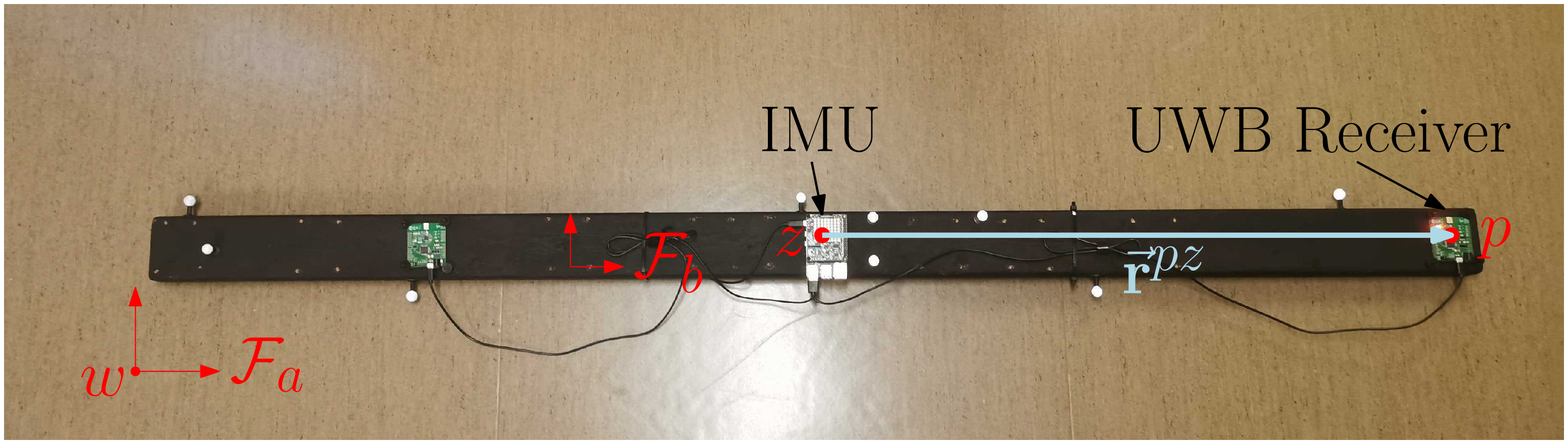}
        \captionof{figure}{The prototype used in simulation and for collection of experimental data. The UWB receiver is used to obtain position measurements from a system of UWB anchors.\vspace{8pt}}
        \label{fig:prototype_cropped}
        \includegraphics[trim=2.2cm 0.1cm 2.2cm -0.2cm, clip=true,width=\textwidth]{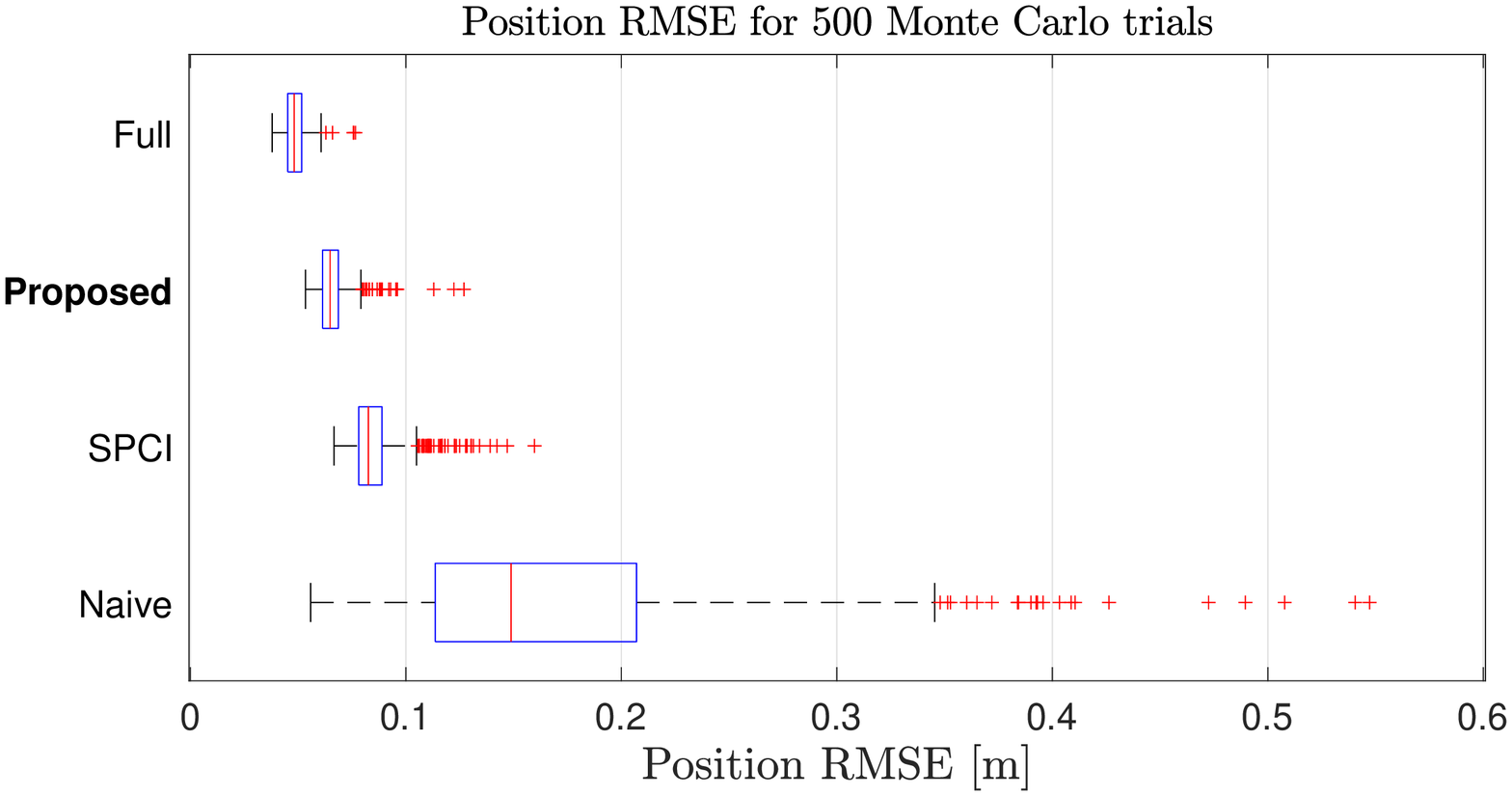}\vspace{-6pt}
		\caption{A box plot showing the median RMSE, outliers, and variation of different estimators over 500 Monte Carlo trials.}
		\label{fig:box_plot}
	\end{minipage}\hspace{10pt}
	\begin{minipage}{0.4\textwidth}
    	\centering
        \footnotesize
        \captionof{table}{RMSE of the estimators averaged 500 trials.}
        \label{tab:mc_rmse}
        \renewcommand{\arraystretch}{1.2}
        \begin{tabular}{|l|c|c|}
            \hline
            & \bfseries Average & \bfseries \% Diff. \\
            & \bfseries RMSE & \bfseries to Full \\
            \hline
            \bfseries Full - Position (m) & 0.0487 & -\\
            \bfseries Proposed - Position (m) & \bfseries 0.0662 & \bfseries 35.9\%\\
            \bfseries SPCI - Position (m) & 0.0862 & 77.0\%\\
            \bfseries Naive - Position (m) & 0.1733 & 256\%\\
            \hline
            \bfseries Full - Attitude (rad) & 0.0190 & -\\
            \bfseries AHRS - Attitude (rad) & 0.0306 & 61.1\%\\
            \hline 
            \end{tabular}
        \normalsize
        \includegraphics[trim=0.9cm 0.1cm 0.5cm -2cm, clip=true,width=\textwidth]{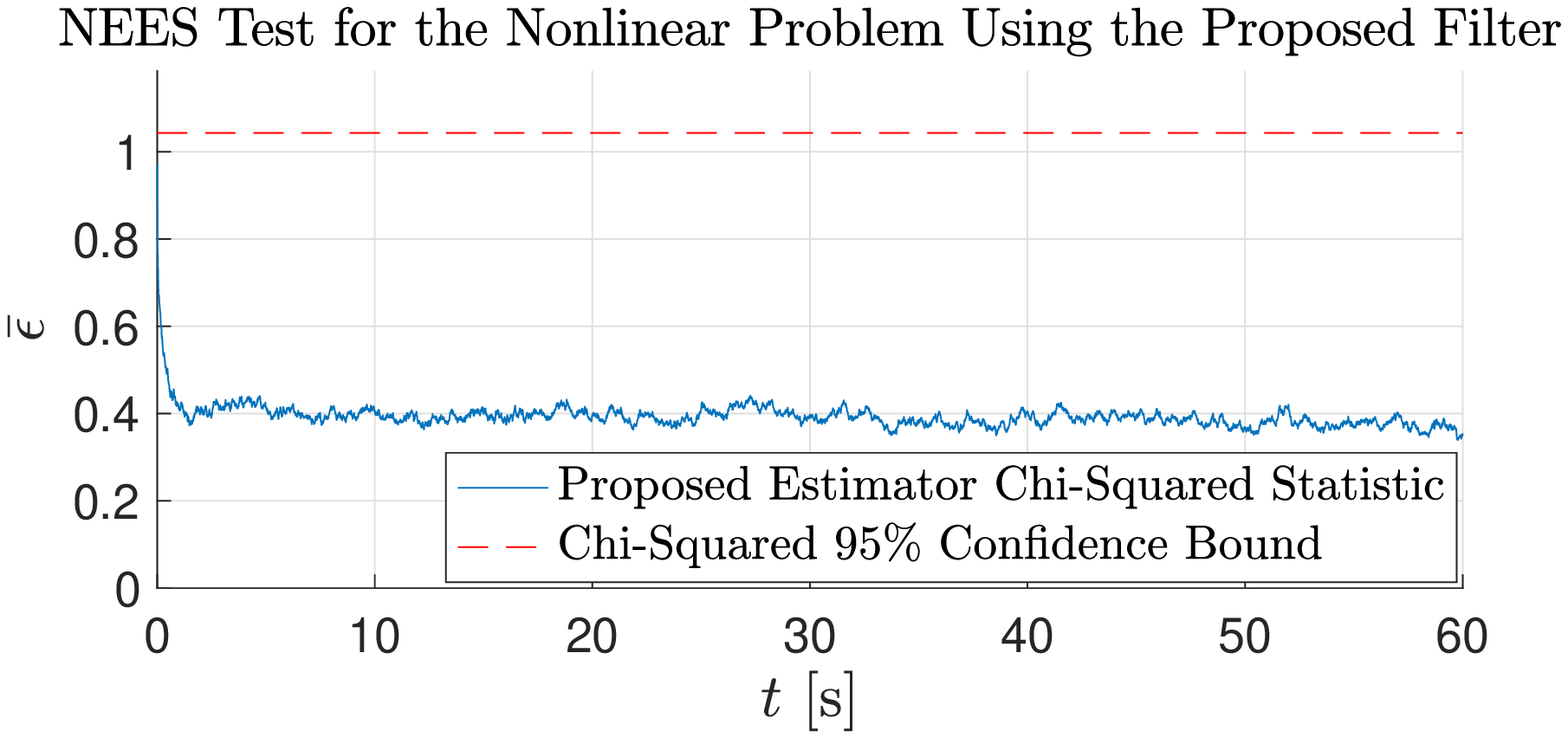}\vspace{-6pt}
    	\captionof{figure}{The NEES test results for the 500 Monte Carlo trials, showing the consistency of the proposed estimator.}
    	\label{fig:mc_nees_nonlinear}
	\end{minipage}
\end{figure*}

To elucidate the benefits of the proposed approach, consider first a toy problem with two discrete-time linear time-invariant processes evolving through
\begin{align}
    x_k^1 &= x_{k-1}^1 - x_{k-1}^2 + w_{k-1}^1, \quad w_{k-1}^1 \sim \mathcal{N} \left(0, q^1\right), \label{eq:sim_linear_process1}\\
    x_k^2 &= x_{k-1}^2 + w_{k-1}^2, \qquad \qquad \hspace{3pt} w_{k-1}^2 \sim \mathcal{N} \left(0, q^2\right), \label{eq:sim_linear_process2}
\end{align}
where $x_k^1 \in \mathbb{R}$ and $x_k^2 \in \mathbb{R}$ are distinct states. Additionally, consider two measurements modelled as 
\begin{align}
    y_k^1 &= x_k^1 + x_k^2 + \nu_k^1, \qquad \nu_{k}^1 \sim \mathcal{N} \left(0, r^1 \right), \label{eq:sim_linear_measurement1}\\
    y_k^2 &= x_k^2 + \nu_k^2, \qquad \qquad \hspace{3pt} \nu_{k}^2 \sim \mathcal{N} \left(0, r^2 \right), \label{eq:sim_linear_measurement2}
\end{align}
where $y_k^1 \in \mathbb{R}$ and $y_k^2 \in \mathbb{R}$ are distinct measurements.

As before, assume at every time-step $k$ the estimates $\hat{x}_{k-1}^1$, $\hat{x}_{k-1}^2$, $\hat{x}_k^2$, the covariances $\hat{P}_{k-1}^1$, $\hat{P}_{k-1}^2$, $\hat{P}_{k}^2$, and the cross-covariance $\hat{P}_{k-1,k-1}^{1,2}$ are known, and the goal is to find the estimate $\hat{x}_k^1$ and its corresponding covariance $\hat{P}_k^1$ using the known values, the process model \eqref{eq:sim_linear_process1}, and the measurement model \eqref{eq:sim_linear_measurement1}. Using \eqref{eq:sim_linear_process2}, $\Psi$ is set to equal $1$, with the definiteness check in place as per Section \ref{subsec:cross_cov_prop}.

Deriving the update equations for $x^1$ yields
\begin{align*}
    \check{x}_k^1 &= \hat{x}_{k-1}^1 - \hat{x}_{k-1}^2, \qquad \hat{x}_k^1 = \check{x}_k^1 + K(y_k^1 - \check{y}_k^1), \\
    \check{P}_k^1 &= \hat{P}_{k-1}^1 + 2 \hat{P}_{k-1,k-1}^{1,2} + \hat{P}_{k-1}^2 + q^1, \\
    \hat{P}_k^1 &= (1-K)^2 \check{P}_k^1 - 2 (1-K)K \check{P}_{k,k}^{1,2} + K^2 (\hat{P}_k^2 + r^1), 
\end{align*}
where the optimal gain $K$ is
\begin{equation}
    K = \left(\check{P}_k^1 + \check{P}_{k,k}^{1,2}\right) \left(\check{P}_k^1 + 2\check{P}_{k,k}^{1,2} + \hat{P}_k^2 + r^1\right)^{-1}. \label{eq:sim_linear_kalman_gain}
\end{equation}
The proposed cascaded filter approximates the cross-covariance terms $\hat{P}_{k-1,k-1}^{1,2}$ and $\check{P}_{k,k}^{1,2}$, and is evaluated against the naive approach in \cite{Lendek2008}, to reiterate the importance of modelling cross-covariances.

To evaluate consistency, 1000 Monte Carlo trials with varying initial conditions and noise realizations are performed. A naive estimator that assumes the estimate $\hat{x}_k^2$ is independent from the estimate $\hat{x}_k^1$ thinks it has access to more information than it actually does, resulting in an overconfident estimator. Therefore, the naive estimator fails both the $\pm 3 \sigma$ test shown in Fig. \ref{fig:SR_linear} and the NEES test in Fig. \ref{fig:MC_NEES_linear}, while the proposed estimator that accommodates for cross-covariances passes both consistency checks. 

The full estimator is clearly the best performer as it can perfectly calculate the cross-covariances and update all states using all measurements. The average RMSE of the proposed framework is 45\% worse than the full estimator, while the naive estimator is 90\% worse. Therefore, this highlights the importance of addressing cross-covariances to prevent the reuse of old information in the filter.

\subsection{Nonlinear System} \label{sec:nonlinear_system}

Consider a rigid body navigating 3D space with an onboard inertial measurement unit (IMU) and an ultra-wideband (UWB) tag, where the position of the UWB tag $p$ relative to the IMU $z$ resolved in the body frame $\mathcal{F}_b$ is known, and is denoted $\mbf{r}_b^{pz} \in \mathbb{R}^3$. Additionally, let there be a $\rho \in \mathbb{N}_{>3}$ number of UWB anchors scattered within ranging distance of the UWB tag, thus providing a noisy position measurement of $\mbf{r}_a^{p_k w} \in\mathbb{R}^3$, which is the position of the UWB tag relative to some arbitrary point $w$ at time-step $k$, in the absolute frame $\mathcal{F}_a$. The simulation parameters of the system are shown in Table~\ref{tab:sim_params_nonlinear}, and the set-up in Fig.~\ref{fig:prototype_cropped}.

Using the gyroscope, the magnetometer, and the accelerometer, an AHRS is designed using the invariant extended Kalman filter (IEKF) \cite{bonnabel2017} framework, where accelerometer aiding follows a thresholding rule similar to the one considered in \cite{farrell2008}. The output of the AHRS is a direction cosine matrix (DCM) estimate $\mbfhat{C}_{ab} \in SO(3)$ that gives the relation $\mbf{r}_a = \mbfhat{C}_{ab} \mbf{r}_{\hat{b}}$, where $\mbf{r}_a$ and $\mbf{r}_{\hat{b}}$ are the same arbitrary vector resolved in $\mathcal{F}_a$, the absolute frame, and $\mathcal{F}_{\hat{b}}$, the estimated body frame, respectively. The corresponding covariance matrix $\mbfhat{P}^\text{AHRS} \in \mathbb{R}^{3 \times 3}$ is also made available by the AHRS.

The position estimator has access to the position measurement $\mbf{y} \in \mathbb{R}^3$ and the accelerometer measurement $\mbf{u}_b \in \mathbb{R}^3$ resolved in the body frame $\mathcal{F}_b$. The state vector of the position estimator is
\begin{equation}
    \mbf{x}(t) = \left[ \begin{array}{c}
        \mbf{r}_a^{zw}(t)  \\
        \mbf{v}_a^{zw}(t) 
    \end{array} \right],
\end{equation}
where $\mbf{r}_a^{zw}(t) \in \mathbb{R}^3$ is the position of the IMU relative to the arbitrary point $w$ resolved in 
$\mathcal{F}_a$, and $\mbf{v}_a^{zw}(t) \in \mathbb{R}^3$ is the velocity of the IMU relative to the point $w$ with respect to $\mathcal{F}_a$, resolved in $\mathcal{F}_a$. The corresponding process model is 
\begin{align}
    \mbfdot{r}_a^{zw}(t) &= \mbf{v}_a^{zw}(t), \label{eq:sim_nonlinear_process} \\
    \mbfdot{v}_a^{zw}(t) &= \mbf{C}_{ab}(t) \left( \mbf{u}_b(t) - \mbf{w}_b(t) \right) + \mbf{g}_a,
\end{align}
where $\mbf{w}_b \in \mathbb{R}^3$ denotes white Gaussian process noise, and $\mbf{g}_a \in \mathbb{R}^3$ is the gravitational acceleration vector resolved in $\mathcal{F}_a$. The discrete-time measurement model is
\begin{equation}
    \mbf{y}_k = \mbf{r}_a^{z_kw} + \mbf{C}_{ab_k} \mbf{r}_b^{p_kz} + \mbs{\nu}_{a_k}, \label{eq:sim_nonlinear_measurement}
\end{equation}
where $\mbs{\nu}_a \in \mathbb{R}^3$ denotes white Gaussian measurement noise. This set-up has inputs in both feeding and receiving filters.

Note that due to the presence of a moment-arm between the IMU and the position sensor, and since the accelerometer measurements are obtained in the body frame $\mathcal{F}_b$, cross-covariances develop between the states of the AHRS and the position estimator. Through knowledge of \eqref{eq:sim_nonlinear_process}-\eqref{eq:sim_nonlinear_measurement}, $\mbf{r}_b^{pz}$, and the output of the AHRS, the goal is to design a consistent position estimator. Four approaches are considered. They are
\begin{enumerate}
    \item an AHRS and a \textbf{naive} sigma point Kalman filter, where cross-covariances are neglected,
    \item an AHRS and the \textbf{proposed} sigma point-based filter, where $\mbshat{\Psi}$ is the identity matrix as per Section \ref{subsec:cross_cov_prop}, 
    \item an AHRS and a sigma point-based Covariance Intersection (\textbf{SPCI}) filter, and
    \item a \textbf{full} sigma point Kalman filter that augments the state vector $\mbf{x}(t)$ with the attitude states $\mbf{C}_{ab}(t)$.
\end{enumerate}

\begin{figure*}
	\centering
	\begin{minipage}{\textwidth}
		\centering
		\includegraphics[trim=0.9cm 0cm 1cm 0cm, clip=true,width=0.33\textwidth]{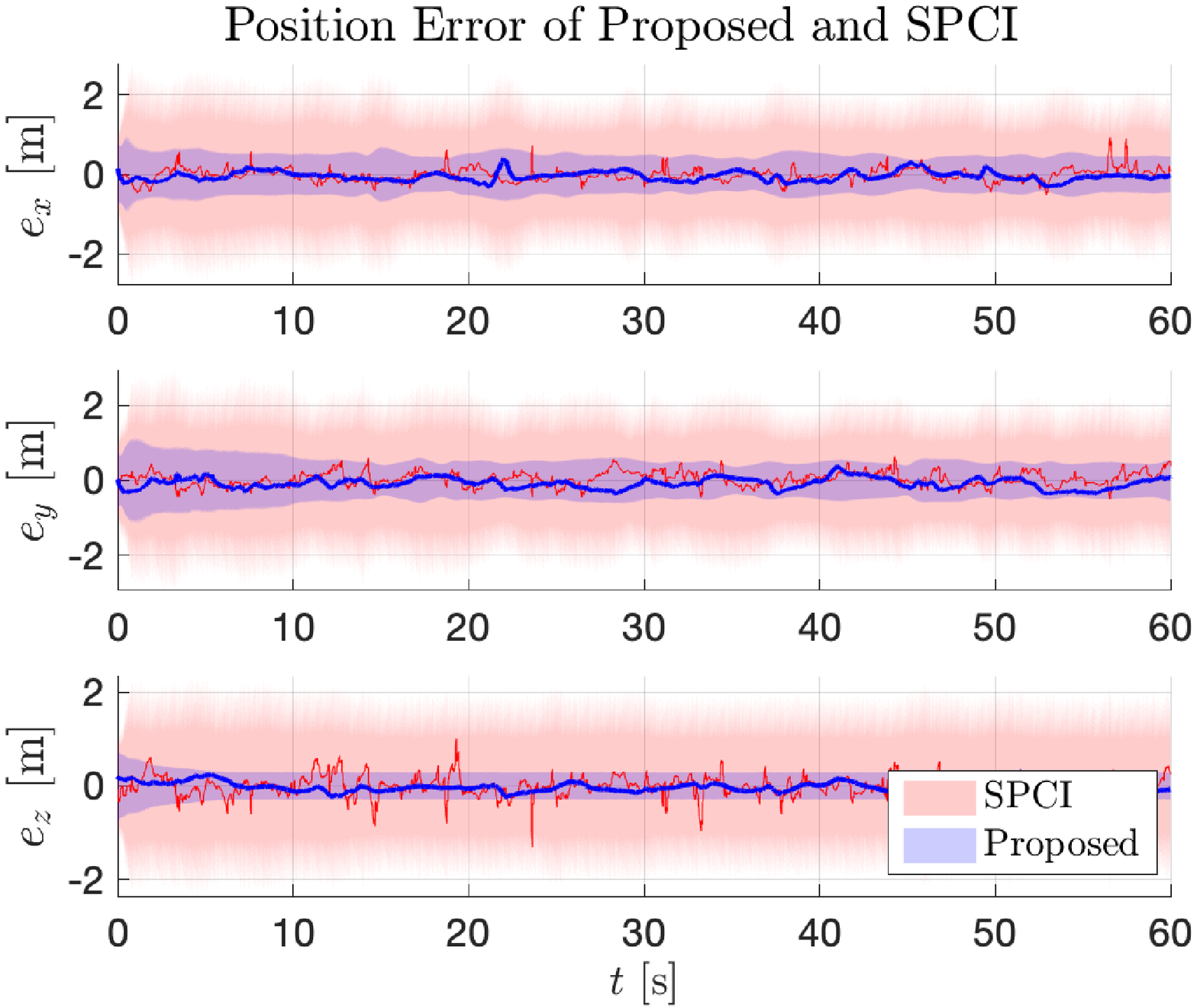}%
		\includegraphics[trim=0.9cm 0cm 1cm 0cm, clip=true,width=0.33\textwidth]{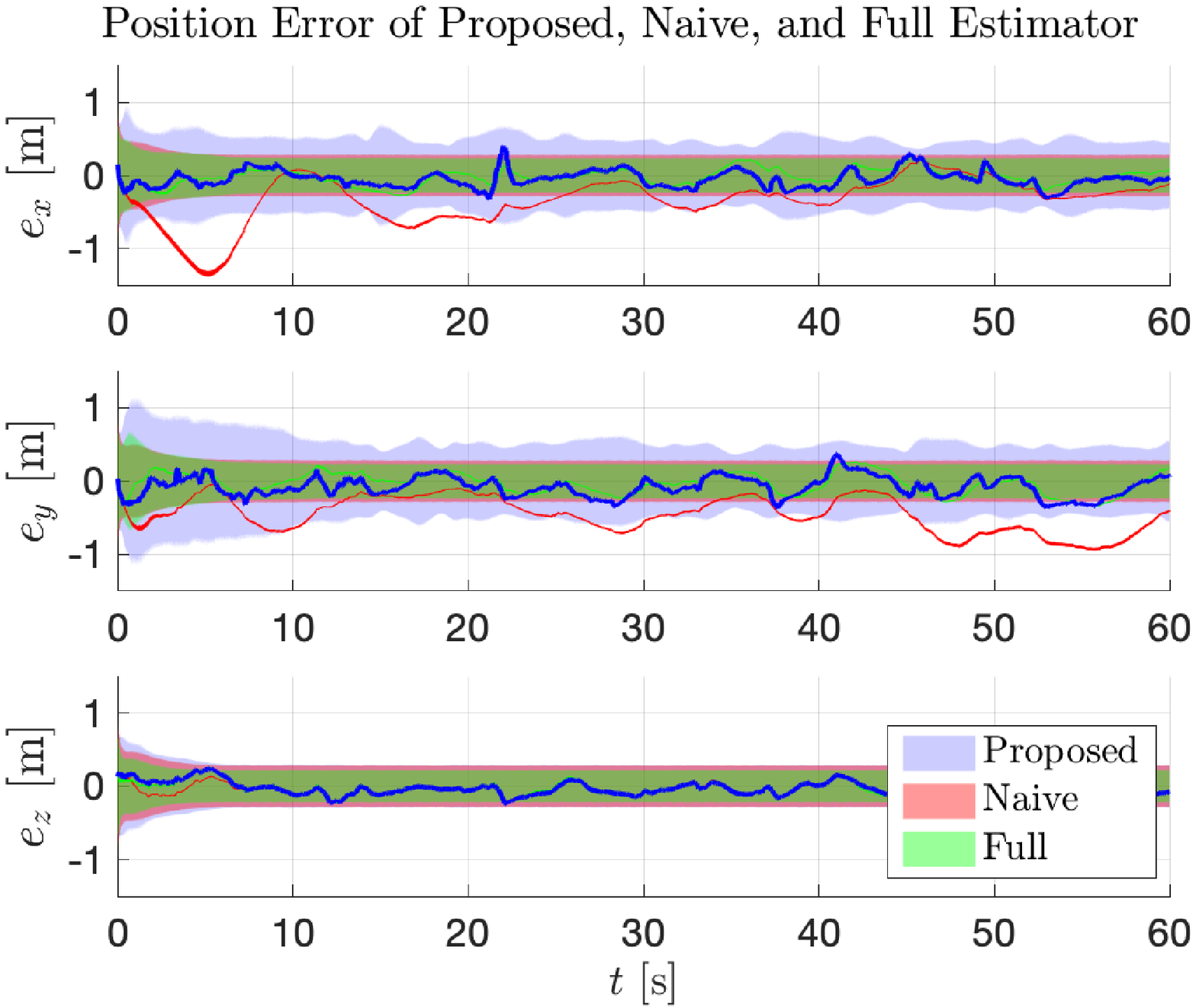}%
		\includegraphics[trim=0.9cm 0cm 1cm 0cm, clip=true,width=0.33\textwidth]{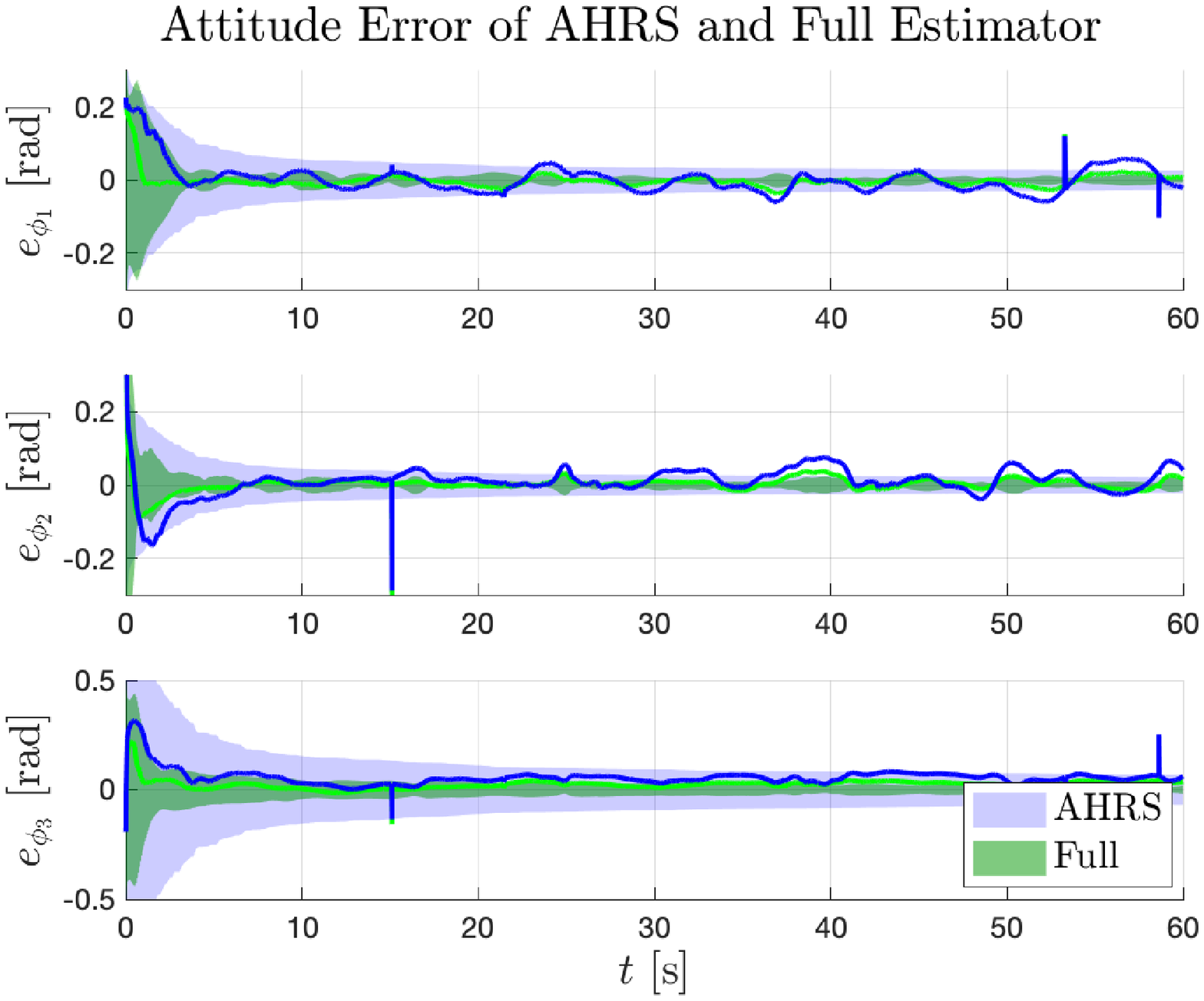}
		\caption{The error trajectories associated with the slow-pace experimental run for the 4 position estimators, and the AHRS. The AHRS is shared by all 3 cascaded estimators. The shaded regions correspond to the $\pm 3 \sigma$ bounds, and the colour of each error trajectory and covariance region are the same.}
		\label{fig:exp_pos_error}
	\end{minipage}
\end{figure*}

The SPCI filter used is a sigma point extension of \cite{Zhu2019}, and is designed in a similar way to the proposed framework. The main difference is that instead of approximating the cross-covariances to compute \eqref{eq:P_v} and \eqref{eq:P_u}, the estimated covariance matrices $\mbfcheck{P}^1$, $\mbfhat{P}^1$, and $\mbfhat{P}^2$ are inflated as per the CI approach, and the cross-covariances are assumed to be zero. This involves tuning an additional scalar parameter. More details on this approach are given in Appendix \ref{appx:spci}.

For the full sigma point Kalman filter, the geodesic $L_2$-mean discussed in \cite{Dai2010} is used for computing means and covariances on $SO(3)$, which are required for sigma point transformations with attitude states. The results of the full sigma point Kalman filter are used as the baseline best possible performance, but in practice would be computationally heavy, inflexible, and even impossible for larger systems.

The computational overhead of the proposed framework is not significantly different from a naive SPKF or the SPCI filter, as all the considered cascaded estimators use sigma point transformations and a similar set of equations. Additionally, a more computationally efficient linearization-based approach is possible as shown in Appendix \ref{appx:linearization}.

To evaluate the 4 estimators, 500 Monte Carlo trials are performed, each 60 seconds long with varying initial conditions and noise realizations. A summary of the position RMSEs is given in Fig. \ref{fig:box_plot} and Table \ref{tab:mc_rmse}. The proposed estimator clearly outperforms both the SPCI and naive estimators, even though all 3 estimators share the same AHRS. As expected, the proposed cascaded filter also passes the consistency test, as shown in Fig. \ref{fig:mc_nees_nonlinear}. 

\begin{table}
\renewcommand{\arraystretch}{1.2}
\caption{RMSE of the different estimators on 3 experimental runs.}
\label{tab:ex_rmse_comp}
\centering
\footnotesize
\begin{tabular}{|l|c|c|c|}
\hline
& \multicolumn{3}{c|}{\bfseries RMSE} \\
\hline
& \bfseries Slow & \bfseries Medium & \bfseries Fast \\
\hline
\bfseries Full - Position (m) & 0.2187 & 0.19613 & 0.17518\\
\bfseries Proposed - Position (m) & \bfseries 0.23138 & \bfseries 0.24818 & \bfseries 0.28261\\
\bfseries SPCI - Position (m) & 0.26690 & 0.32152 & 0.32045\\
\bfseries Naive - Position (m) & 0.69616 & 0.50807 & 1.3655\\
\hline
\bfseries Full - Attitude (rad) & 0.04400 & 0.03225 & 0.02896\\
\bfseries AHRS - Attitude (rad) & 0.08934 & 0.07669 & 0.13757\\
\hline
\end{tabular}
\normalsize
\end{table}

The proposed framework provides a 35.9\% worse position estimate than the full estimator even though the provided attitude estimates are 61.1\% worse. The worse attitude estimates are due to the UWB measurements correcting the attitude states in the full estimator, but not in cascaded architectures as discussed in Section \ref{sec:prob_formulation}. However, the loss of performance is compensated by the modularity, computational gain, and flexibility of the proposed cascaded filter.

\section{Experimental Results} \label{sec:experimental}

Experimental data is collected for the nonlinear system discussed in Section \ref{sec:nonlinear_system} using the prototype shown in Fig.~\ref{fig:prototype_cropped}. The IMU data is collected using a Raspberry Pi Sense HAT at 240 Hz, and the position measurements are collected at 16 Hz using the Pozyx Creator Kit, which is a UWB-based positioning system. Five UWB anchors communicate with a UWB tag placed on the body 84 cm away from the IMU. This is complemented with ground truth data collected using an OptiTrack optical motion capture system at 120 Hz. 

Three datasets are tested using the four estimators discussed in Section \ref{sec:nonlinear_system}. Each run involves moving the rigid body randomly in a volume of approximately 5 m $\times$ 4 m $\times$ 2 m while recording the IMU, UWB, and ground truth data. The main difference between the three datasets is the pace at which the robot is moved around and rotated. 

\begin{figure}
    \centering
    \includegraphics[trim=0.9cm 0cm 1cm 0cm, clip=true,width=\columnwidth]{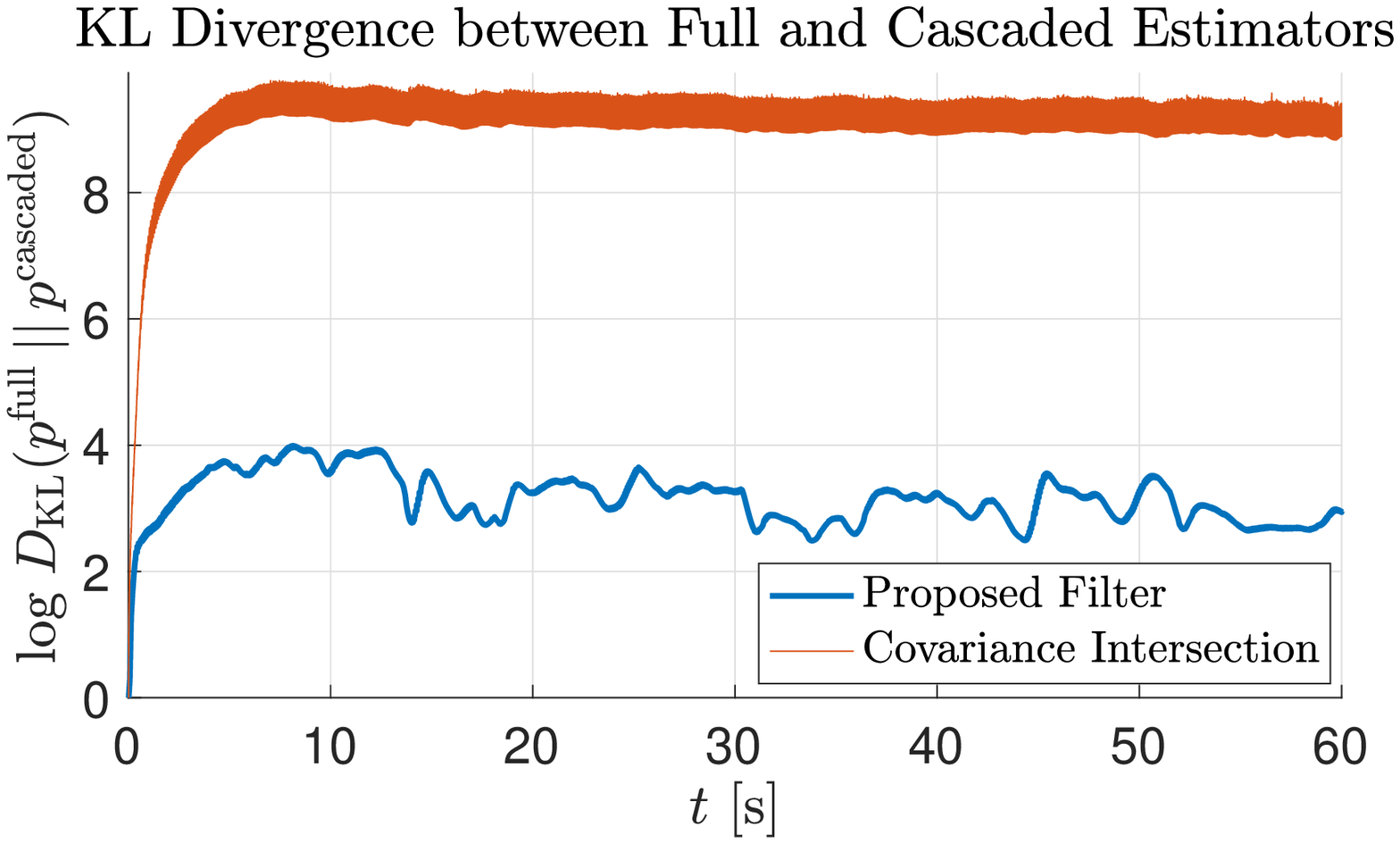}
    \caption{The logarithm of the KL divergence measure associated with the slow-pace experimental run. The KL divergence is computed between the estimated distribution of each cascaded estimator and the full estimator.}
		\label{fig:KL_slow}
\end{figure}

The $\pm 3 \sigma $ bound error plots for the 4 estimators on the slow pace experiment are shown in Fig. \ref{fig:exp_pos_error}, and an RMSE-based comparison on the three datasets is shown in Table \ref{tab:ex_rmse_comp}. As expected, the naive cascaded filter performs poorly, while the performance of the proposed cascaded filter beats the SPCI approach and is comparable to the full estimator. Not only does the proposed framework achieve a lower position RMSE, but it is also less conservative as compared to the SPCI. This is further displayed in Fig. \ref{fig:KL_slow}, where the KL divergence \cite[Chapter 9]{Kullback1968} measure shows that the estimated distribution of the proposed estimator is closer to that of the full estimator, which is the best available estimate of the true distribution. 

Due to imperfectly calibrated hardware, further unmodelled phenomena such as indoor magnetic perturbations, and nonlinearities of the system, the AHRS in the slow-pace experiment takes over 15 seconds to achieve near steady-state conditions, as shown in Fig. \ref{fig:exp_pos_error}. Additionally, the AHRS performs poorly particularly in the fast pace experiment, as shown in Table \ref{tab:ex_rmse_comp}. However, the proposed framework still achieves a performance comparable to the full estimator, which shows that the approximation discussed in Section \ref{subsec:cross_cov_prop} is reasonable, even for such complex systems. Additionally, the worse attitude estimates as compared to the simulation runs means that the cross-covariances between the AHRS and the position estimator are more significant, which is why the naive estimator performs so poorly. This emphasizes the significance of the proposed estimator, particularly when estimation error is persistent in the feeding filter.

\section{Conclusion} \label{sec:conclusion}

In this letter, the importance of modelling cross-covariances in cascaded estimation is addressed, and a novel approach is introduced using a sigma point transformation. The main contribution is the probabilistic approximation of the propagation of cross-covariance terms when the local filters do not estimate a common set of states. The proposed approach is compared to the naive filtering approach and a Covariance Intersection filter approach both in simulation and in experiment. Superior, consistent results are achieved for both a linear and a nonlinear problem. Comparable performance to a full non-cascaded estimator is also achieved, but with all the advantages of cascaded filtering such as improved flexibility and reduced computational complexity.

\section*{Acknowledgments}

The authors would like to thank Natalia Pavlasek for her role in collecting the experimental data.

\appendices

\section{} \label{appx:deriv_meas_update}

The conditional joint distribution of $\mbf{x}_k^1, \mbf{x}_k^2, \mbf{y}_k^1 \big \vert \mathcal{I}_{k-1}^1, \mathcal{I}_{k}^2$ is assumed to be Gaussian according to
\begin{equation*}
     \mathcal{N} \left( \left[ \begin{array}{c}
        \mbfcheck{x}_k^1  \\
         \mbfhat{x}_k^2 \\
         \mbfcheck{y}_k^1
    \end{array} \right], \left[ \begin{array}{ccc}
        \mbfcheck{P}_k^1 & \mbfcheck{P}_{k,k}^{1,2} & \mbs{\Sigma}_{\mbfcheck{x}^1_k \mbfcheck{y}^1_k} \\
        \big(\mbfcheck{P}_{k,k}^{1,2}\big)^\trans & \mbfhat{P}_k^2 & \mbs{\Sigma}_{\mbfhat{x}^2_k \mbfcheck{y}^1_k} \\
        \mbs{\Sigma}_{\mbfcheck{x}^1_k \mbfcheck{y}^1_k}^\trans & \mbs{\Sigma}_{\mbfhat{x}^2_k \mbfcheck{y}^1_k}^\trans & \mbs{\Sigma}_{\mbfcheck{y}^1_k \mbfcheck{y}^1_k}
    \end{array} \right] \right).
\end{equation*}
Using \cite[Lemma A.2]{sarkka2013} to condition on $\mbf{y}_k^1$ as well, the conditional joint distribution of $\mbf{x}_k^1, \mbf{x}_k^2 \big \vert \mathcal{I}_{k}^1, \mathcal{I}_{k}^2$ is given by 
\footnotesize
\begin{align*}
    \mathcal{N} \Bigg( &\Bigg[ \begin{array}{c}
        \mbfcheck{x}^1_k + \mbs{\Sigma}_{\mbfcheck{x}^1_k \mbfcheck{y}^1_k}\mbs{\Sigma}_{\mbfcheck{y}^1_k \mbfcheck{y}^1_k}^{-1}\left(\mbf{y}_k^1 - \mbfcheck{y}^1_k\right)  \\
         \mbfhat{x}^2_k + \mbs{\Sigma}_{\mbfhat{x}^2_k \mbfcheck{y}^1_k}\mbs{\Sigma}_{\mbfcheck{y}^1_k \mbfcheck{y}^1_k}^{-1}\left(\mbf{y}_k^1 - \mbfcheck{y}^1_k\right)
    \end{array} \Bigg],\\
    &\Bigg[ \begin{array}{cc}
        \mbfcheck{P}_k^1 - \mbs{\Sigma}_{\mbfcheck{x}^1_k \mbfcheck{y}^1_k}\mbs{\Sigma}_{\mbfcheck{y}^1_k \mbfcheck{y}^1_k}^{-1}\mbs{\Sigma}_{\mbfcheck{x}^1_k \mbfcheck{y}^1_k}^\trans & \mbfcheck{P}_{k,k}^{1,2}- \mbs{\Sigma}_{\mbfcheck{x}^1_k \mbfcheck{y}^1_k}\mbs{\Sigma}_{\mbfcheck{y}^1_k \mbfcheck{y}^1_k}^{-1}\mbs{\Sigma}_{\mbfhat{x}^2_k \mbfcheck{y}^1_k}^\trans \\
        \big(\mbfcheck{P}_{k,k}^{1,2}\big)^\trans- \mbs{\Sigma}_{\mbfhat{x}^2_k \mbfcheck{y}^1_k}\mbs{\Sigma}_{\mbfcheck{y}^1_k \mbfcheck{y}^1_k}^{-1}\mbs{\Sigma}_{\mbfcheck{x}^1_k \mbfcheck{y}^1_k}^\trans & \mbfhat{P}_k^2- \mbs{\Sigma}_{\mbfhat{x}^2_k \mbfcheck{y}^1_k}\mbs{\Sigma}_{\mbfcheck{y}^1_k \mbfcheck{y}^1_k}^{-1}\mbs{\Sigma}_{\mbfhat{x}^2_k \mbfcheck{y}^1_k}^\trans
    \end{array} \Bigg] \Bigg).
\end{align*}
\normalsize
Using \cite[Lemma A.2]{sarkka2013} again, and by replacing the conditioning on $\mathcal{I}_{k}^2$ using $\mbfhat{x}_{k}^2 \big( \mathcal{I}_{k}^2 \big)$ and $\mbfhat{P}_{k}^2 \big( \mathcal{I}_{k}^2 \big)$ as was done in Section \ref{sec:prob_formulation}, the distribution of  $\mbf{x}_k^1 \vert \mathcal{I}_{k}^1, \mbfhat{x}_{k}^2, \mbfhat{P}_{k}^2$ is given by
\footnotesize
\begin{align*}
    \mathcal{N} \Big( &\mbfcheck{x}^1_k + \mbs{\Sigma}_{\mbfcheck{x}^1_k \mbfcheck{y}^1_k}\mbs{\Sigma}_{\mbfcheck{y}^1_k \mbfcheck{y}^1_k}^{-1}\Big(\mbf{y}_k^1 - \mbfcheck{\mbf{y}}^1_k\Big) +  \Big(\mbfcheck{P}_{k,k}^{1,2}- \mbs{\Sigma}_{\mbfcheck{x}^1_k \mbfcheck{y}^1_k}\mbs{\Sigma}_{\mbfcheck{y}^1_k \mbfcheck{y}^1_k}^{-1}\mbs{\Sigma}_{\mbfhat{x}^2_k \mbfcheck{y}^1_k}^\trans\Big)
    \nonumber\\&\Big(\mbfhat{P}_k^2- \mbs{\Sigma}_{\mbfhat{x}^2_k \mbfcheck{y}^1_k}\mbs{\Sigma}_{\mbfcheck{y}^1_k \mbfcheck{y}^1_k}^{-1}\mbs{\Sigma}_{\mbfhat{x}^2_k \mbfcheck{y}^1_k}^\trans\Big)^{-1}\mbs{\Sigma}_{\mbfhat{x}^2_k \mbfcheck{y}^1_k}\mbs{\Sigma}_{\mbfcheck{y}^1_k \mbfcheck{y}^1_k}^{-1}\Big(\mbf{y}_k^1 - \mbfcheck{y}^1_k\Big), \nonumber\\
    &\mbfcheck{P}_k^1 - \mbs{\Sigma}_{\mbfcheck{x}^1_k \mbfcheck{y}^1_k}\mbs{\Sigma}_{\mbfcheck{y}^1_k \mbfcheck{y}^1_k}^{-1}\mbs{\Sigma}_{\mbfcheck{x}^1_k \mbfcheck{y}^1_k}^\trans - \Big(\mbfcheck{P}_{k,k}^{1,2}- \mbs{\Sigma}_{\mbfcheck{x}^1_k \mbfcheck{y}^1_k}\mbs{\Sigma}_{\mbfcheck{y}^1_k \mbfcheck{y}^1_k}^{-1}\mbs{\Sigma}_{\mbfhat{x}^2_k \mbfcheck{y}^1_k}^\trans\Big) \nonumber\\
    &\Big( \mbfhat{P}_k^2- \mbs{\Sigma}_{\mbfhat{x}^2_k \mbfcheck{y}^1_k}\mbs{\Sigma}_{\mbfcheck{y}^1_k \mbfcheck{y}^1_k}^{-1}\mbs{\Sigma}_{\mbfhat{x}^2_k \mbfcheck{y}^1_k}^\trans \Big)^{-1}\Big( \big(\mbfcheck{P}_{k,k}^{1,2}\big)^\trans- \mbs{\Sigma}_{\mbfhat{x}^2_k \mbfcheck{y}^1_k}\mbs{\Sigma}_{\mbfcheck{y}^1_k \mbfcheck{y}^1_k}^{-1}\mbs{\Sigma}_{\mbfcheck{x}^1_k \mbfcheck{y}^1_k}^\trans \Big)\Big).
\end{align*}
 \normalsize
Therefore, the filter equations using a Bayesian approach are given by \eqref{eq:x1hatSimplified}-\eqref{eq:k12simplified}.

\section{Sigma Point-Based Covariance Intersection} \label{appx:spci}

\begingroup
\allowdisplaybreaks

Consider the problem of fusing two estimates
\begin{align}
    \mbf{x}^1 &\sim \mathcal{N} \left( \mbfhat{x}^1, \mbfhat{P}^1 \right), \\
    \mbf{x}^2 &\sim \mathcal{N} \left( \mbfhat{x}^2, \mbfhat{P}^2 \right),
\end{align}
of the same state vector $\mbf{x}$, where the cross-covariance matrix 
\begin{equation}
	\mbf{P}^{1,2} = \mathbb{E} \left[ \left(\mbf{x} - \mbf{x}^1\right)		\left(\mbf{x} - \mbf{x}^2\right)^\trans  \right]
\end{equation} 
is unknown. One consistent way to obtain a state estimate $\mbfhat{x}$ with a covariance matrix $\mbfhat{P}$ by fusing $\mbf{x}^1$ and $\mbf{x}^2$ is the Covariance Intersection (CI) method \cite{Julier1997, Uhlmann2003}. The core of this approach is to disregard any cross-covariances by inflating the joint covariance matrices $\mbfhat{P}^1$ and $\mbfhat{P}^2$. Therefore, the assumed joint distribution between $\mbf{x}^1$ and $\mbf{x}^2$ is given by
\begin{equation}
	\left[ \begin{array}{c}
		\mbf{x}^1 \\
		\mbf{x}^2
	\end{array}	 \right] \sim \mathcal{N} \left(\left[ \begin{array}{c}
		\mbfhat{x}^1 \\
		\mbfhat{x}^2
	\end{array}	 \right] , \left[ \begin{array}{cc}
		\f{1}{w}\mbfhat{P}^1 & \mbf{0} \\
		\mbf{0} & \f{1}{1-w}\mbfhat{P}^2
	\end{array}	 \right] \right),
\end{equation}
where $w \in \mathbb{R}$ is a weighting parameter.

\begin{figure}
    \centering
    \includegraphics[width=0.85\columnwidth]{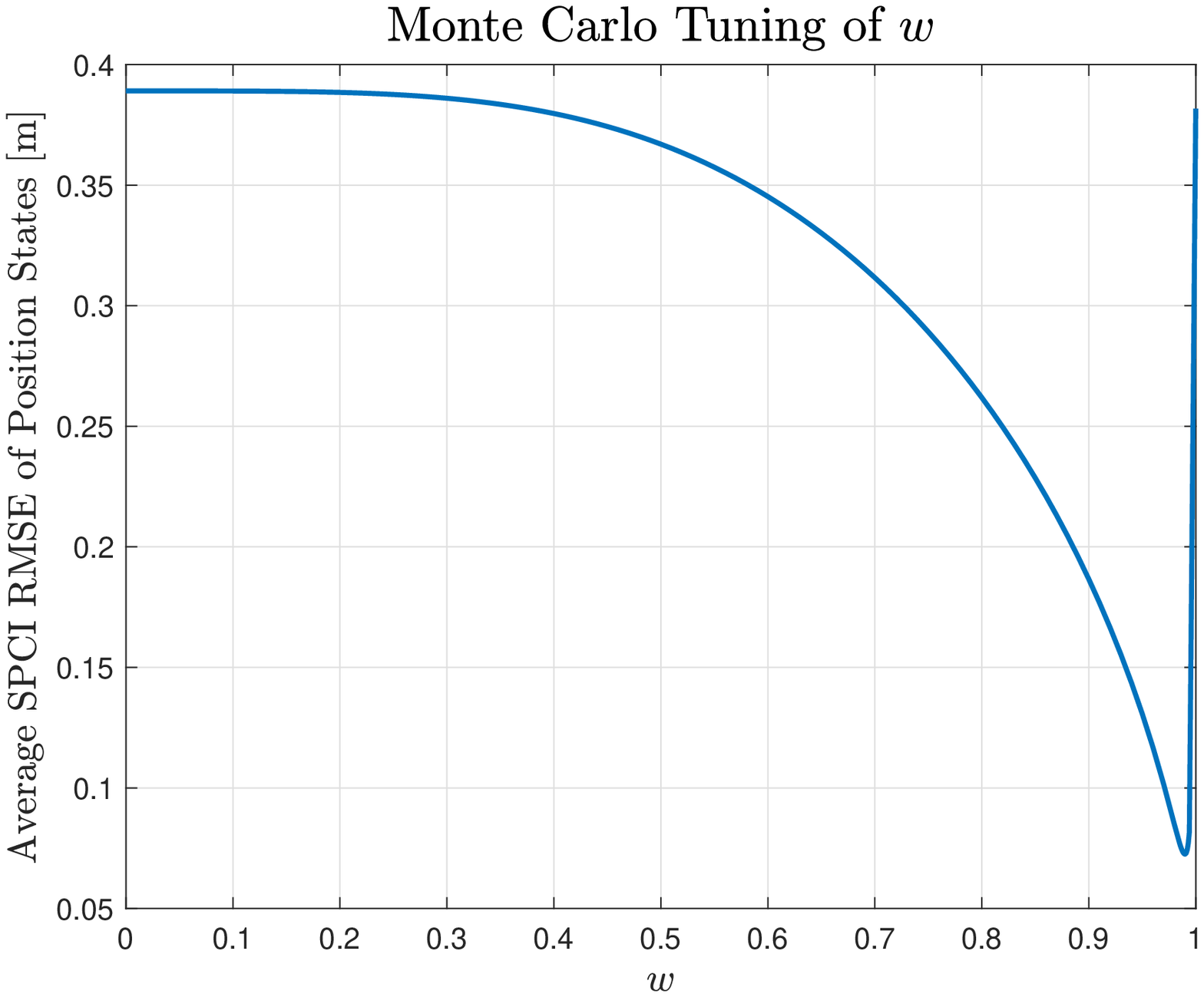}
    \caption{The RSME of the SPCI filter averaged over 50 different trajectories for different values of $w$, showing best performance at $w=0.99$.}
    \label{fig:ci_tuning}
\end{figure}

In Sections \ref{sec:sim} and \ref{sec:experimental}, a sigma point Covariance Intersection (SPCI) approach is used as an alternative, less accurate solution to the proposed framework. The proposed framework attempts to approximate the cross-covariance matrices, while CI inflates the joint covariances matrices to avoid doing so. Therefore, designing a SPCI approach is similar to the proposed framework. The state vectors $\mbf{x}_{k-1}^1$ and $\mbf{x}_{k-1}^2$ and the process noise $\mbf{w}_{k-1}^1$ are augmented into one vector as in \eqref{eq:v_augment}. The difference, however, is in the covariance matrix, which now is of the form
\begin{align}
    \mbfhat{P}_{\mbf{v}_{k-1}} &= \left[ \begin{array}{ccc}
        \f{1}{w}\mbfhat{P}_{k-1}^1 & \mbf{0} & \mbf{0} \\
        \mbf{0} & \f{1}{1-w}\mbfhat{P}_{k-1}^2 & \mbf{0} \\
        \mbf{0} & \mbf{0} & \mbf{Q}_{k-1}^1
    \end{array} \right]
\end{align}
instead of \eqref{eq:P_v}.
Similarly, for the correction step, the new augmented state vector is of the form \eqref{eq:P_u} and the new covariance matrix is of the form
\begin{equation}
        \mbfhat{P}_{\mbf{u}_k}= \left[ \begin{array}{ccc}
            \f{1}{w}\mbfcheck{P}_{k}^1 & \mbf{0} & \mbf{0} \\
            \mbf{0} & \f{1}{1-w}\mbfcheck{P}_{k}^2 & \mbf{0} \\
            \mbf{0} & \mbf{0} & \mbf{R}_{k}^1
        \end{array} \right].
\end{equation}

One of the disadvantages of the Covariance Intersection method is that it requires tuning of an additional scalar parameter, $w$. Usually, when the two vectors being fused are different estimates of the same state vector, the weighting factors are the result of an optimization approach that minimizes the trace or some other metric of the fused covariance matrix. However, in the problem of cascaded filtering when state vectors being estimated are distinct, the different units of the diagonal elements of the fused covariance matrix would mean this approach yields biased results. Another approach involves choosing the weighting parameters and the Kalman gain to minimize the trace of the posterior covariance matrix, as in \cite{Zhu2019}. However, it is unclear how this approach could be utilized in a sigma point-based approach. Therefore, the value of the weighting parameter $w$ is tuned using Monte Carlo trials as shown in Fig. \ref{fig:ci_tuning}, leading to the choice of $w=0.99$. 

\section{Linearization-Based Approach} \label{appx:linearization}

\subsection{Proposed Linearization-Based Cascaded Receiving Filter}

In this section, a linearization-based approach is derived using the proposed framework. Deriving a linearized model from a nonlinear system is standard, using a first-order Taylor series approximation. Therefore, rather than considering a nonlinear system and addressing its linearized counterpart, a discrete-time linear system of the form 
\begin{align*}
	\mbf{x}_{k}^1 &= \mbf{A}_{k-1}^1 \mbf{x}_{k-1}^1 + \mbf{B}_{k-1}^1 \mbf{x}_{k-1}^2 + \mbf{L}_{k-1}^1 \mbf{w}_{k-1}^1,\\
    \mbf{y}_{k}^1 &= \mbf{C}_k^1 \mbf{x}_k^1 + \mbf{D}_k^1 \mbf{x}_k^2 + \mbf{M}_k^1 \mbs{\nu}_k^1, \\
    \mbf{x}_{k}^2 &= \mbf{A}_{k-1}^2 \mbf{x}_{k-1}^2 + \mbf{L}_{k-1}^2 \mbf{w}_{k-1}^2, \\
    \mbf{y}_{k}^2 &= \mbf{C}_k^2 \mbf{x}_k^2 + \mbf{M}_k^2 \mbs{\nu}_k^2,
\end{align*}
will be considered for conciseness of notation, where
\begin{align*}
\mbf{w}_{k-1}^1 &\sim \mathcal{N}\left( \mbf{0}, \mbf{Q}^1_{k-1} \right), \quad
\mbs{\nu}_{k}^1 \sim \mathcal{N}\left( \mbf{0}, \mbf{R}^1_{k} \right), \\
\mbf{w}_{k-2}^1 &\sim \mathcal{N}\left( \mbf{0}, \mbf{Q}^2_{k-1} \right), \quad
\mbs{\nu}_{k}^2 \sim \mathcal{N}\left( \mbf{0}, \mbf{R}^2_{k} \right).
\end{align*}
In what follows, the notation
    $\mathcal{I}_{k} = \mathcal{I}_{k}^1 \cup \mathcal{I}_{k}^2$
is used.

\subsubsection{Prediction Step}

When propagating from time-step $k-1$ to $k$, the state estimates $\mbfhat{x}_{k-1}^1$, $\mbfhat{x}_{k-1}^2$ and the covariance matrices $\mbfhat{P}_{k-1}^1$, $\mbfhat{P}_{k-1}^2$, $\mbfhat{P}_{k-1,k-1}^{1,2}$ are known. Therefore, the prediction step is straightforward, and is given by
    \begin{align*}
        \mbfcheck{x}_k^1 &= \mbf{A}_{k-1}^1 \mbfhat{x}_{k-1}^1 + \mbf{B}_{k-1}^1 \mbfhat{x}_{k-1}^2, \\
        \mbfcheck{P}_{k}^1 &= \mathbb{E}\left[ \left( \mbf{x}_k^1 - \mbfcheck{x}_k^1 \right)\left( \mbf{x}_k^1 - \mbfcheck{x}_k^1 \right)^\trans \Big\vert \hspace{3pt} \mathcal{I}_{k-1} \right] \\
        &= \mathbb{E}\Big[ \big( \mbf{A}_{k-1}^1\mbf{x}_{k-1}^1 + \mbf{B}_{k-1}^1\mbf{x}_{k-1}^2 + \mbf{L}_{k-1}^1\mbf{w}_{k-1}^1 \\&\hspace{23pt}- \mbf{A}_{k-1}^1\mbfhat{x}_{k-1}^1 + \mbf{B}_{k-1}^1\mbfhat{x}_{k-1}^2 \big)\left( \cdot \right)^\trans \Big\vert \hspace{3pt} \mathcal{I}_{k-1} \Big] \\
        &= \mathbb{E}\Big[ \big( \mbf{A}_{k-1}^1 \left(\mbf{x}_{k-1}^1 - \mbfhat{x}_{k-1}^1 \right)+ \mbf{B}_{k-1}^1 \left(\mbf{x}_{k-1}^2 - \mbfhat{x}_{k-1}^2 \right) \\&\hspace{23pt}+ \mbf{L}_{k-1}^1\mbf{w}_{k-1}^1 \big)\left( \cdot \right)^\trans \Big\vert \hspace{3pt} \mathcal{I}_{k-1} \Big] \\
        &= \mbf{A}_{k-1}^1\mbfhat{P}_{k-1}^1\left(\mbf{A}_{k-1}^1\right)^\trans + \mbf{A}_{k-1}^1 \mbfhat{P}_{k-1,k-1}^{1,2} \left(\mbf{B}_{k-1}^1\right)^\trans \\&\hspace{12pt}+ \mbf{B}_{k-1}^1 \left( \mbfhat{P}_{k-1,k-1}^{1,2} \right)^\trans\left(\mbf{A}_{k-1}^1\right)^\trans \\&\hspace{12pt}+ \mbf{B}_{k-1}^1 \mbfhat{P}_{k-1}^2 \left( \mbf{B}_{k-1}^1 \right)^\trans + \mbf{L}_{k-1}^1 \mbf{Q}_{k-1}^1 \left(\mbf{L}_{k-1}^1\right)^\trans, \\
        \mbfcheck{P}_{k,k-1}^{1,2} &= \mathbb{E}\left[ \left( \mbf{x}_k^1 - \mbfcheck{x}_k^1 \right)\left( \mbf{x}_{k-1}^2 - \mbfhat{x}_{k-1}^2 \right)^\trans \big\vert \hspace{3pt} \mathcal{I}_{k-1} \right] \\
        &= \mathbb{E}\Big[ \big( \mbf{A}_{k-1}^1\mbf{x}_{k-1}^1 + \mbf{B}_{k-1}^1\mbf{x}_{k-1}^2 + \mbf{L}_{k-1}^1\mbf{w}_{k-1}^1 \\&\hspace{10pt}- \mbf{A}_{k-1}^1\mbfhat{x}_{k-1}^1 + \mbf{B}_{k-1}^1\mbfhat{x}_{k-1}^2 \big)\big( \mbf{x}_{k-1}^2 - \mbfhat{x}_{k-1}^2 \Big)^\trans \big\vert \hspace{3pt} \mathcal{I}_{k-1} \Big] \\
        &= \mathbb{E}\Big[ \big( \mbf{A}_{k-1}^1 \left(\mbf{x}_{k-1}^1 - \mbfhat{x}_{k-1}^1 \right)+ \mbf{B}_{k-1}^1 \left(\mbf{x}_{k-1}^2 - \mbfhat{x}_{k-1}^2 \right) \\&\hspace{23pt}+ \mbf{L}_{k-1}^1\mbf{w}_{k-1}^1 \big)\left( \mbf{x}_{k-1}^2 - \mbfhat{x}_{k-1}^2 \right)^\trans \Big\vert \hspace{3pt} \mathcal{I}_{k-1} \Big] \\
        &= \mbf{A}_{k-1}^1 \mbfhat{P}_{k-1,k-1}^{1,2} + \mbf{B}_{k-1}^1 \mbfhat{P}_{k-1}^2,
    \end{align*}
    where $(\operatorname{expression})(\cdot)^\trans$ is used to denote $(\operatorname{expression})(\operatorname{expression})^\trans$, and the assumptions 
    \begin{align*}
        \mathbb{E} \Big[ \left( \mbf{x}_{k-1}^1 - \mbfhat{x}_{k-1}^1 \right)\left(\mbf{w}_{k-1}^1\right)^\trans \big\vert \hspace{3pt} \mathcal{I}_{k-1} \Big] &= \mbf{0}, \\
        \mathbb{E} \Big[ \left( \mbf{x}_{k-1}^2 - \mbfhat{x}_{k-1}^2 \right) \left(\mbf{w}_{k-1}^1\right)^\trans \big\vert \hspace{3pt} \mathcal{I}_{k-1} \Big] &= \mbf{0}
    \end{align*}
    are made. Finding $\mbfcheck{P}_{k,k}^{1,2}$ from $\mbfcheck{P}_{k,k-1}^{1,2}$ then follows the approximation given in Section \ref{subsec:cross_cov_prop}. 

\subsubsection{Correction Step}

When correcting the predicted state at time-step $k$ using measurements $\mbf{y}_k^1$, the state estimates $\mbfcheck{x}_k^1$, $\mbfhat{x}_k^2$ and the covariance matrices $\mbfcheck{P}_{k}^1$, $\mbfhat{P}_k^2$, $\mbfcheck{P}_{k,k}^{1,2}$ are known. The posterior distribution of the states of the receiving filter, conditioned on the states of the feeding filter is derived in Appendix \ref{appx:deriv_meas_update} and Section \ref{sec:prob_formulation}. Based on this, a filter of the form \eqref{eq:x1hatSimplified}-\eqref{eq:k12simplified} is given. The analytical covariance terms $\mbs{\Sigma}_{\mbfcheck{x}^1_k \mbfcheck{y}^1_k}$, $\mbs{\Sigma}_{\mbfhat{x}^2_k \mbfcheck{y}^1_k}$, $\mbs{\Sigma}_{\mbfcheck{y}^1_k \mbfcheck{y}^1_k}$ for the linear system are then derived to be
\begin{align*}
    \mbs{\Sigma}_{\mbfcheck{x}^1_k \mbfcheck{y}^1_k} &= \mathbb{E} \left[ \left( \mbf{x}_k^1 - \mbfcheck{x}_k^1 \right)\left( \mbf{y}_k^1 - \mbfcheck{y}_k^1 \right)^\trans \big\vert \hspace{3pt} \mathcal{I}_{k} \right] \\
    &= \mathbb{E} \Big[ \big(\mbf{x}_k^1 - \mbfcheck{x}_k^1\big) \big( \mbf{C}_k^1 \mbf{x}_k^1 + \mbf{D}_k^1 \mbf{x}_k^2 + \mbf{M} \mbs{\nu}_k^1 \\&\hspace{75pt}- \mbf{C}_k^1 \mbfcheck{x}_k^1 - \mbf{D}_k^1 \mbfhat{x}_k^2 \big\vert \hspace{3pt} \mathcal{I}_{k} \big)^\trans \Big] \\
    &= \mathbb{E} \Big[ \big(\mbf{x}_k^1 - \mbfcheck{x}_k^1\big) \big( \mbf{C}_k^1 \left( \mbf{x}_k^1 - \mbfcheck{x}_k^1 \right) + \\&\hspace{75pt}\mbf{D}_k^1 \left( \mbf{x}_k^2 - \mbfhat{x}_k^2 \right) + \mbf{M} \mbs{\nu}_k^1 \big)^\trans \big\vert \hspace{3pt} \mathcal{I}_{k} \Big] \\
    &= \mbfcheck{P}_k^1 \left(\mbf{C}_k^1\right)^\trans + \mbfcheck{P}_{k,k}^{1,2} \left(\mbf{D}_k^1\right)^\trans, \\
    \mbs{\Sigma}_{\mbfhat{x}^2_k \mbfcheck{y}^1_k} &= \mathbb{E} \left[ \left( \mbf{x}_k^2 - \mbfhat{x}_k^2 \right)\left( \mbf{y}_k^1 - \mbfcheck{y}_k^1 \right)^\trans \big\vert \hspace{3pt} \mathcal{I}_{k} \right] \\
    &= \mathbb{E} \Big[ \big(\mbf{x}_k^2 - \mbfhat{x}_k^2\big) \big( \mbf{C}_k^1 \mbf{x}_k^1 + \mbf{D}_k^1 \mbf{x}_k^2 + \mbf{M} \mbs{\nu}_k^1 \\&\hspace{75pt}- \mbf{C}_k^1 \mbfcheck{x}_k^1 - \mbf{D}_k^1 \mbfhat{x}_k^2 \big)^\trans \big\vert \hspace{3pt} \mathcal{I}_{k} \Big] \\
    &= \mathbb{E} \Big[ \big(\mbf{x}_k^2 - \mbfhat{x}_k^2\big) \big( \mbf{C}_k^1 \left( \mbf{x}_k^1 - \mbfcheck{x}_k^1 \right) + \\&\hspace{75pt}\mbf{D}_k^1 \left( \mbf{x}_k^2 - \mbfhat{x}_k^2 \right) + \mbf{M} \mbs{\nu}_k^1 \big)^\trans \big\vert \hspace{3pt} \mathcal{I}_{k} \Big] \\
    &= \left(\mbfcheck{P}_{k,k}^{1,2}\right)^\trans \left(\mbf{C}_k^1\right)^\trans + \mbfhat{P}_{k}^{2} \left(\mbf{D}_k^1\right)^\trans, \\
    \mbs{\Sigma}_{\mbfcheck{y}^1_k \mbfcheck{y}^1_k} &= \mathbb{E} \left[ \left( \mbf{y}_k^1 - \mbfcheck{y}_k^1 \right)\left( \mbf{y}_k^1 - \mbfcheck{y}_k^1 \right)^\trans \big\vert \hspace{3pt} \mathcal{I}_{k} \right] \\
    &= \mathbb{E} \Big[ \big( \mbf{C}_k^1 \mbf{x}_k^1 + \mbf{D}_k^1 \mbf{x}_k^2 + \mbf{M} \mbs{\nu}_k^1 - \mbf{C}_k^1 \mbfcheck{x}_k^1 - \mbf{D}_k^1 \mbfhat{x}_k^2 \big) \left( \cdot \right)^\trans \big\vert \hspace{3pt} \mathcal{I}_{k} \Big] \\
    &= \mathbb{E} \Big[ \big( \mbf{C}_k^1 \left( \mbf{x}_k^1 - \mbfcheck{x}_k^1 \right) + \mbf{D}_k^1 \left( \mbf{x}_k^2 - \mbfhat{x}_k^2 \right) + \mbf{M} \mbs{\nu}_k^1 \big) \left( \cdot \right)^\trans \big\vert \hspace{3pt} \mathcal{I}_{k} \Big] \\
    &= \mbf{C}_k^1 \mbfcheck{P}_k^1 \left(\mbf{C}_k^1\right)^\trans + \mbf{C}_k^1 \mbfcheck{P}_{k,k}^{1,2} \left(\mbf{D}_k^1\right)^\trans + \mbf{D}_k^1 \left(\mbfcheck{P}_{k,k}^{1,2}\right)^\trans \left(\mbf{C}_k^1\right)^\trans \\&\hspace{86pt}+ \mbf{D}_k^1 \mbfhat{P}_k^2 \left(\mbf{D}_k^1\right)^\trans + \mbf{M}_k^1 \mbf{R}_k^1 \left(\mbf{M}_k^1\right)^\trans,
\end{align*}

\begin{figure*}
	\begin{minipage}{0.54\textwidth}
        	\centering
        	\includegraphics[trim=0.7cm 0cm 1cm 0cm, clip=true,width=\textwidth]{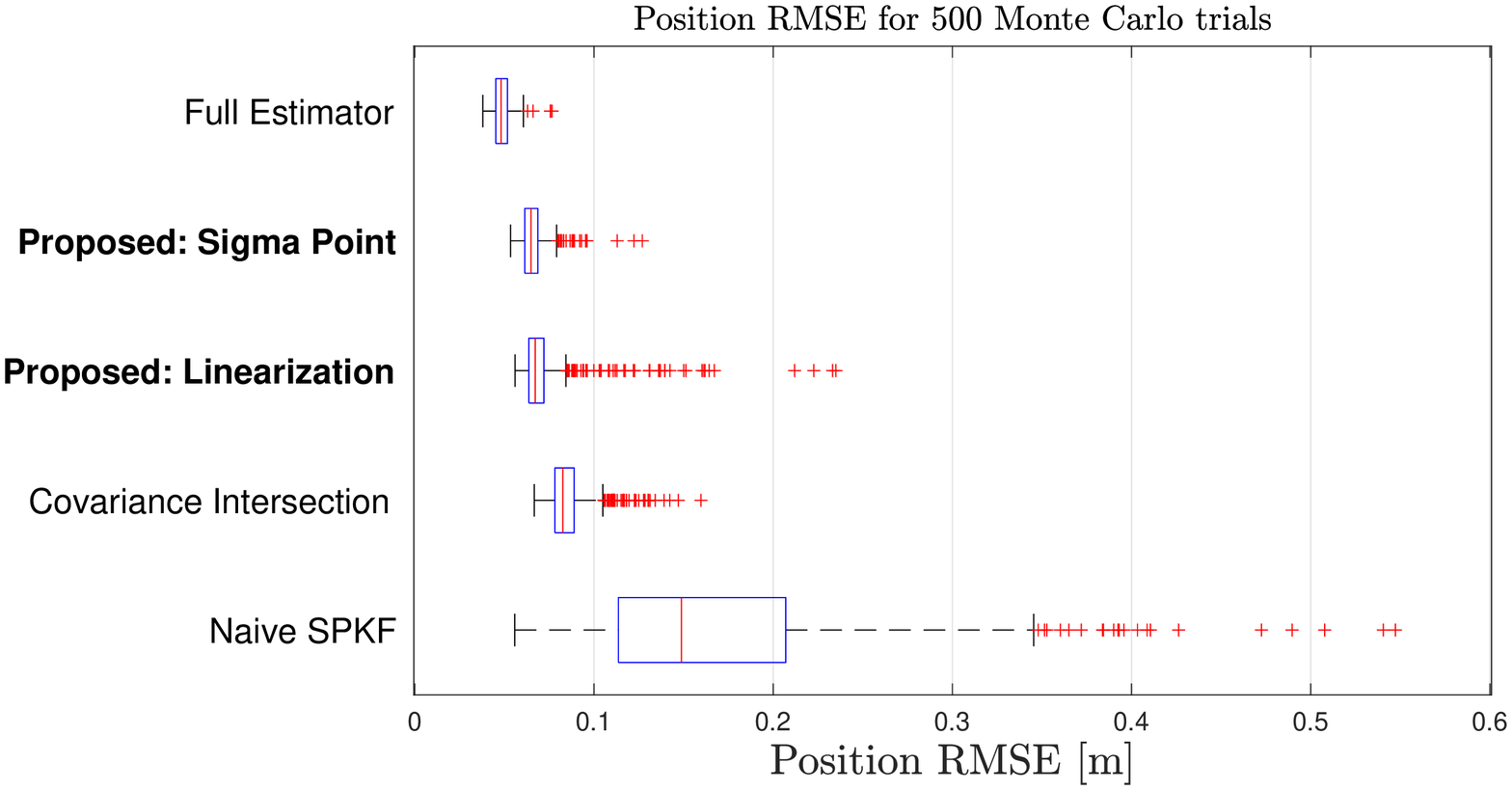}%
        	\caption{A box plot showing the median RMSE, outliers, and variation over 500 Monte Carlo trials.\vspace{10pt}} 
        	\label{fig:box_appx}
        	\includegraphics[width=\textwidth]{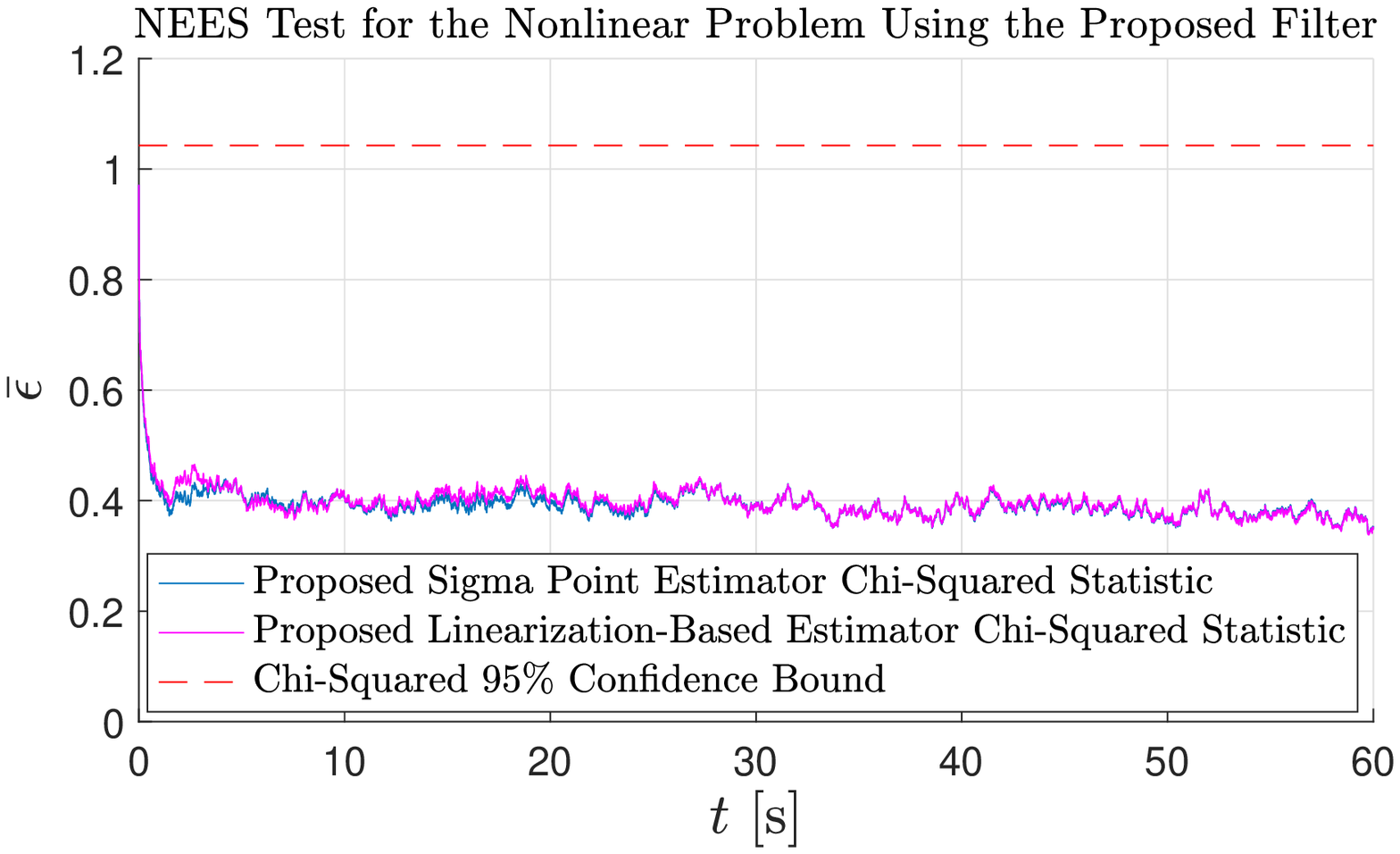}
            \captionof{figure}{The NEES test results for the 500 Monte Carlo trials, showing the consistency of both proposed estimators.\vspace{10pt}}
            \label{fig:nees_appx}
	\end{minipage}\hspace{10pt}
	\begin{minipage}{0.43\textwidth}
        \centering
        \footnotesize
        \captionof{table}{RMSE of the estimators averaged 500 trials.}
        \label{tab:mc_rmse_appx}
        \begin{tabular}{|l|c|c|}
            \hline
            & \bfseries Average & \bfseries \% Diff. \\
            & \bfseries RMSE & \bfseries to Full \\
            \hline
            \bfseries Full - Position (m) & 0.0487 & -\\
            \bfseries Proposed: SP - Position (m) & \bfseries 0.0662 & \bfseries 35.9\%\\
            \bfseries Proposed: Lin. - Position (m) & \bfseries 0.0729 & \bfseries 49.7\% \\
            \bfseries SPCI - Position (m) & 0.0862 & 77.0\%\\
            \bfseries Naive - Position (m) & 0.1733 & 256\%\\
            \hline
            \bfseries Full - Attitude (rad) & 0.0190 & -\\
            \bfseries AHRS - Attitude (rad) & 0.0306 & 61.1\%\\
            \hline 
        \end{tabular}
        \includegraphics[trim=0cm 0cm 0cm -1.8cm,clip=true,width=\textwidth]{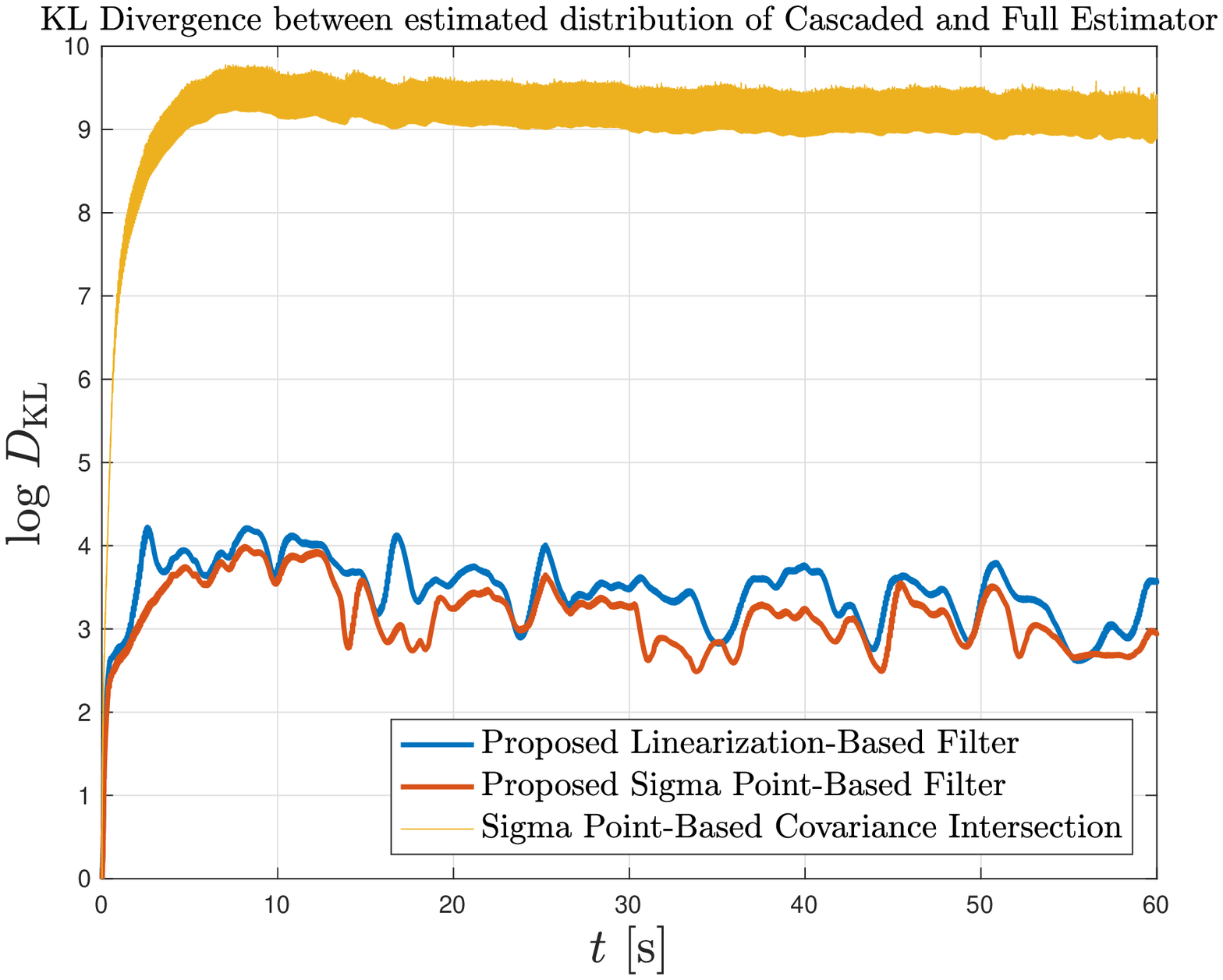}
        \captionof{figure}{The logarithm of the KL divergence measure associated with the experimental run at a slow pace. The KL divergence is computed between each cascaded estimator and the full estimator.}
        \label{fig:kl_appx}
	\end{minipage}
\end{figure*}

where $(\operatorname{expression})(\cdot)^\trans$ is used to denote $(\operatorname{expression})(\operatorname{expression})^\trans$, and the assumptions 
\begin{align*}
    \mathbb{E} \Big[ \left( \mbf{x}_k^1 - \mbfcheck{x}_k^1 \right) \left(\mbs{\nu}_{k}^1\right)^\trans \big\vert \hspace{3pt} \mathcal{I}_{k} \Big] &= \mbf{0}, \\
    \mathbb{E} \Big[ \left( \mbf{x}_k^2 - \mbfhat{x}_k^2 \right) \left(\mbs{\nu}_{k}^1\right)^\trans \big\vert \hspace{3pt} \mathcal{I}_{k} \Big] &= \mbf{0}
\end{align*}
are made.

\subsection{Simulation Results}

The nonlinear problem addressed in Section \ref{sec:sim} will also be used here to evaluate the linearization-based approach. To evaluate the different approaches, 500 Monte Carlo trials with varying initial conditions and noise realizations are performed as in Section \ref{sec:sim}. A summary of the results are given in Fig. \ref{fig:box_appx} and Table \ref{tab:mc_rmse_appx}. The proposed sigma point estimator achieves the best performance and beats both the linearization-based approach and the SPCI. The linearization-based approach on average performs better than the SPCI estimator. However, the presence of more significant outliers is possibly due to linearization errors, while the SPCI approach considered here uses sigma points and does not suffer from linearization errors. Even then, the SPCI is still on average outperformed by the significantly less computationally complex linearization-based approach. A linearization-based CI approach is possible, but is not considered as further approximations associated with linearization are expected to provide worse results anyway.

A NEES test with 5\% significance level is also performed to evaluate the consistency of the proposed estimators. As seen in Fig. \ref{fig:nees_appx}, and as per the theory validating the NEES test, the hypothesis that the estimator is consistent cannot be rejected with 95\% confidence.

\subsection{Experimental Results}

\begin{figure*}
    \begin{minipage}{\textwidth}
		\centering
		\vspace{10pt}
		\includegraphics[trim=0.9cm 0cm 1cm 0cm, clip=true,width=0.33\textwidth]{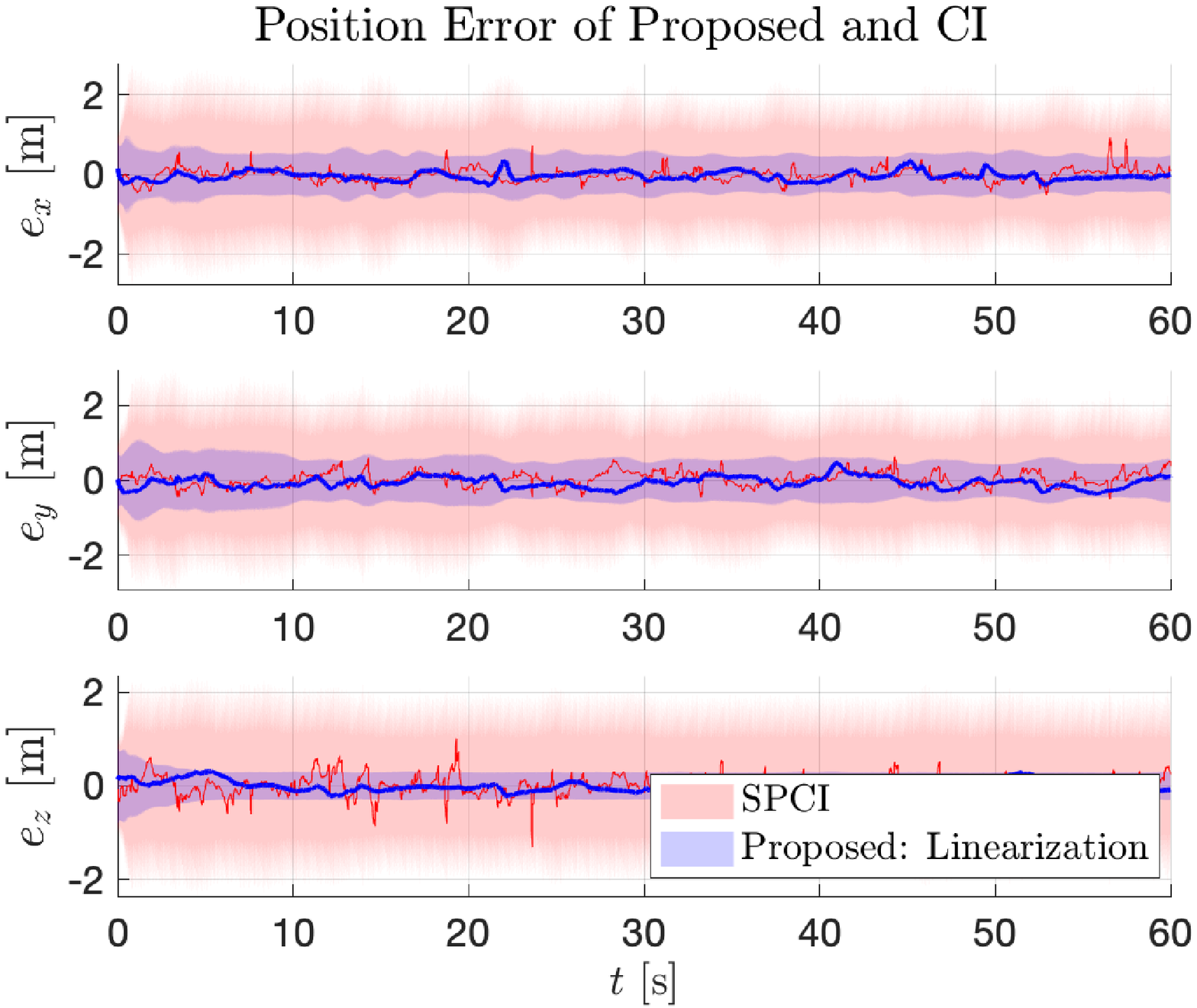}%
		\includegraphics[trim=0.9cm 0cm 1cm 0cm, clip=true,width=0.33\textwidth]{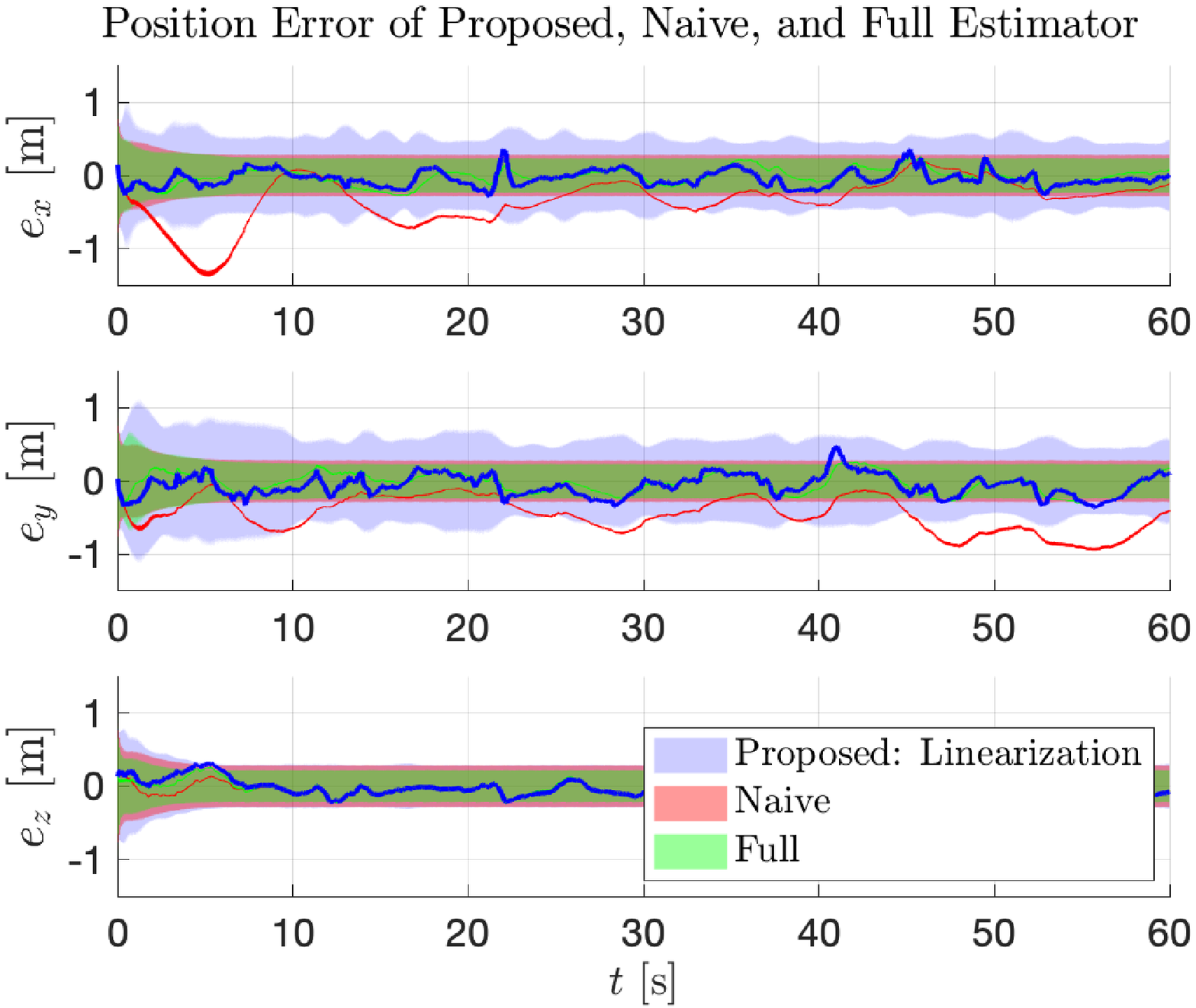}
		\includegraphics[trim=0.9cm 0cm 1cm 0cm, clip=true,width=0.33\textwidth]{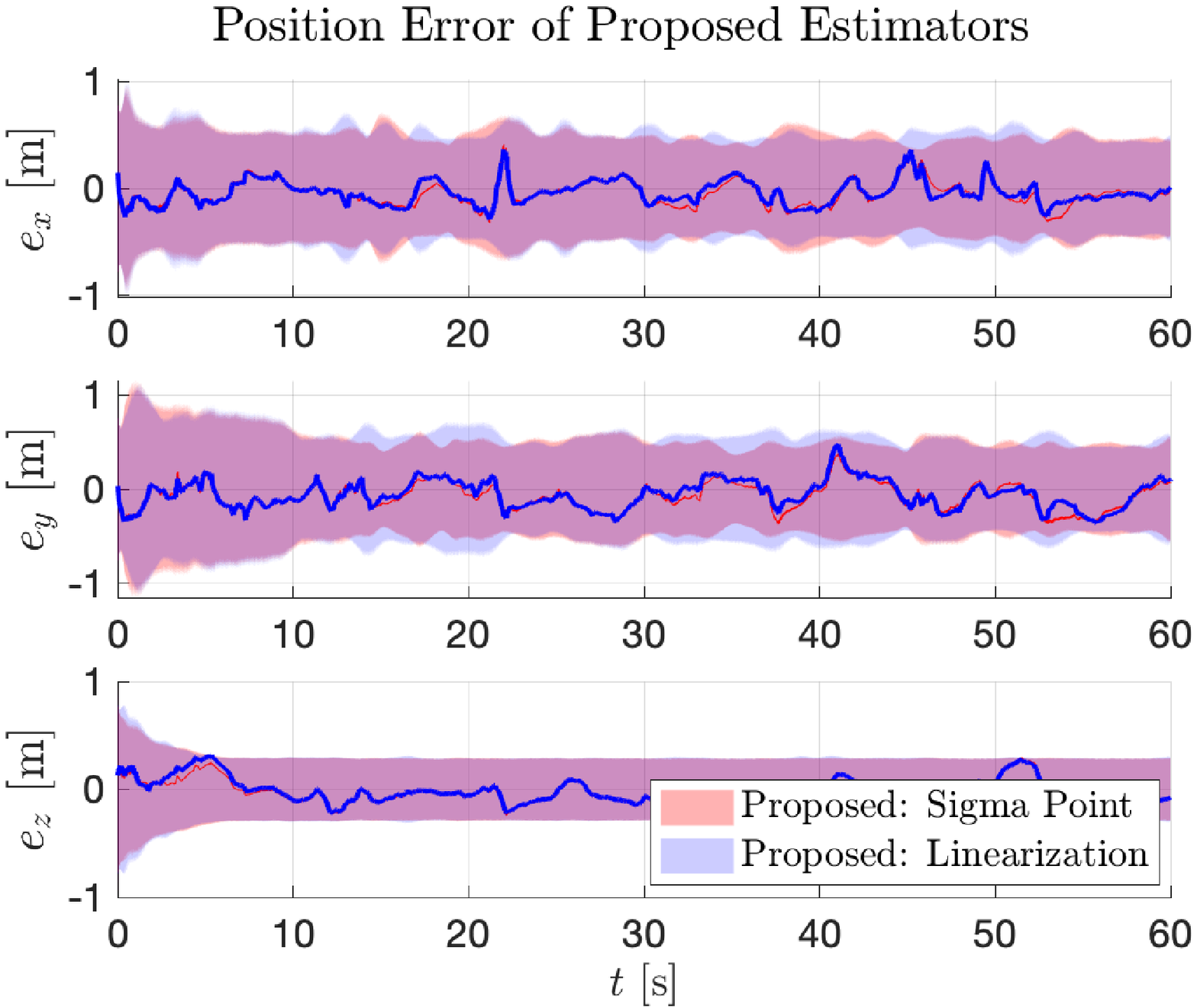}
		\caption{The position errors associated with the experimental run at a slow pace. The shaded region corresponds to the $\pm3\sigma$ bound, and the colour of each error trajectory and covariance region are the same.}
		\label{fig:pos_error_exp_appx}
	\end{minipage}
\end{figure*}

The linearization-based approach is also tested on the slow-paced experimental run. The error trajectory plots and corresponding $\pm3\sigma$ bounds are shown in Fig. \ref{fig:pos_error_exp_appx}. The proposed linearization-based approach behaves similarly to the proposed sigma point approach, and as is the case with the sigma point approach, it remains mostly within the bounds while being less over-conservative than the SPCI approach. This can also be seen in Fig. \ref{fig:kl_appx}, where the KL divergence \cite[Chapter 9]{Kullback1968} measure shows that the estimated distribution of the linearization-based approach is much closer to the estimated distribution of the full estimator when compared to the SPCI, which is the best available estimate of the true distribution. 

The position RMSE of the proposed linearization-based approach is 0.23494 m, compared to the proposed sigma point-based estimator's of 0.23138 m and the full estimator's of 0.21870 m. Both proposed approaches are similar to the performance of the full estimator, while the linearization-based approach is slightly worse as linearization errors contribute to a slightly less accurate estimated distribution, as shown in Fig. \ref{fig:kl_appx}. 

\ifCLASSOPTIONcaptionsoff
  \newpage
\fi

\bibliographystyle{IEEEtran}
\bibliography{IEEEabrv,cf_ral_extended}

\end{document}